\documentclass{article} % For LaTeX2e
\usepackage[preprint]{colm2026_conference}

\usepackage{microtype}
\usepackage{hyperref}
\usepackage{url}
\usepackage{booktabs}
\usepackage{graphicx}
\usepackage{subfig}

\usepackage{lineno}

\definecolor{darkblue}{rgb}{0, 0, 0.5}
\hypersetup{colorlinks=true, citecolor=darkblue, linkcolor=darkblue, urlcolor=darkblue}

\title{Synthetic Users, Real Differences: an Evaluation Framework for User Simulation in Multi-Turn Conversations}

\author{%
Yu Lu Liu$^{1*}$ \quad Hyokun Yun \quad Tanya Roosta$^2$ \quad Ziang Xiao$^1$\\
$^1$Johns Hopkins University\quad $^2$UC Berkeley\\
\texttt{\{yliu624, ziang.xiao\}@jhu.edu}\\
\texttt{yungilbert@gmail.com}\\
\texttt{tanya55@berkeley.edu}\\
}

\newcommand{\para}[1]{\vspace{2pt}\noindent\textbf{#1}~}
\newcommand\FrameworkName{{\mbox{\fontfamily{qcr}\selectfont realsim}}}
\begin{document}

% \ifcolmsubmission
% \linenumbers
% \fi

\maketitle

\begin{abstract}
There is growing interest in exploring user simulation as an alternative to gathering and scoring real user-chatbot interactions for AI chatbot evaluation. For this purpose, it is important to ensure the realism of the simulation, i.e., the extent to which simulated dialogues reflect real dialogues users have with chatbots. Most existing methods evaluating simulation realism produce coarse quality signal and remain solely at the level of individual dialogues. To support more rigorous evaluation in this area, we propose \FrameworkName, an evaluation framework that enables practitioners to take a distributional view of real vs. simulated dialogues along 8 dimensions, covering attributes related to the communicative functions of the interaction, user states, and the surface form of user messages. We then instantiate the framework with a curated dataset of 1K multi-turn task-focused real user-chatbot dialogues that cover 16 domains of chatbot applications.
Overall, we find that simulated users tend to struggle at capturing communication frictions that real users introduce to interactions, which could make evaluations based on such simulations overly optimistic. 
We also observe variability in performance across different domains, which may indicate a need for domain-specific user simulators. %
\end{abstract}

\section{Introduction}

The evaluation of large language model (LLM)-based chatbots is a critical problem given their increasing popularity, application and impact on society. Their open-ended and complex output space, as well as the multi-turn nature of the interaction, make this problem a particularly challenging one. Given that human evaluation or user studies can be limiting in terms of efficiency and reproducibility, there is growing interest in using LLMs to simulate users and generate synthetic interactions, which can be then used to evaluate the chatbot’s quality  \cite[e.g.,][]{IQA-EVAL, ChatBench, MINT}. 
The validity of such methods depends on the realism of the user simulation: if simulated interactions are not representative of the ways real human users interact with chatbots, any resulting evaluation outcomes would struggle to capture chatbot performance in real-world setting. 

Current meta-evaluation methods often present human annotators or LLM-as-a-judge with simulated dialogues one by one, requesting them to judge how realistic each one seems to be. This approach tend to produce a coarse quality signal (e.g., realistic vs. not realistic) rather than highlighting specific aspects that make some dialogues seem less realistic than others. Moreover, making judgment of realism solely at the level of individual dialogues may fail to identify distributional patterns that distinguish simulated users from real users. 

To address these gaps, we propose \FrameworkName, an evaluation framework that assesses the realism of user simulation methods by comparing real vs. simulated dialogues along 8 dimensions, at 3 levels of user behaviors: \textit{communicative functions} of the interaction (user intent, feedback), \textit{user states} (user emotion, domain-specific knowledge, personal context), and the \textit{surface form} of user messages (message length, linguistic attributes, errors). We instantiate the framework with a curated dataset of 1K multi-turn task-focused real user-chatbot dialogues that cover 16 domains of chatbot applications, allowing us to gather domain-specific insights about the behaviors of simulated users. 

\begin{figure*}[t!]
    \centering
    \includegraphics[scale = 0.75]{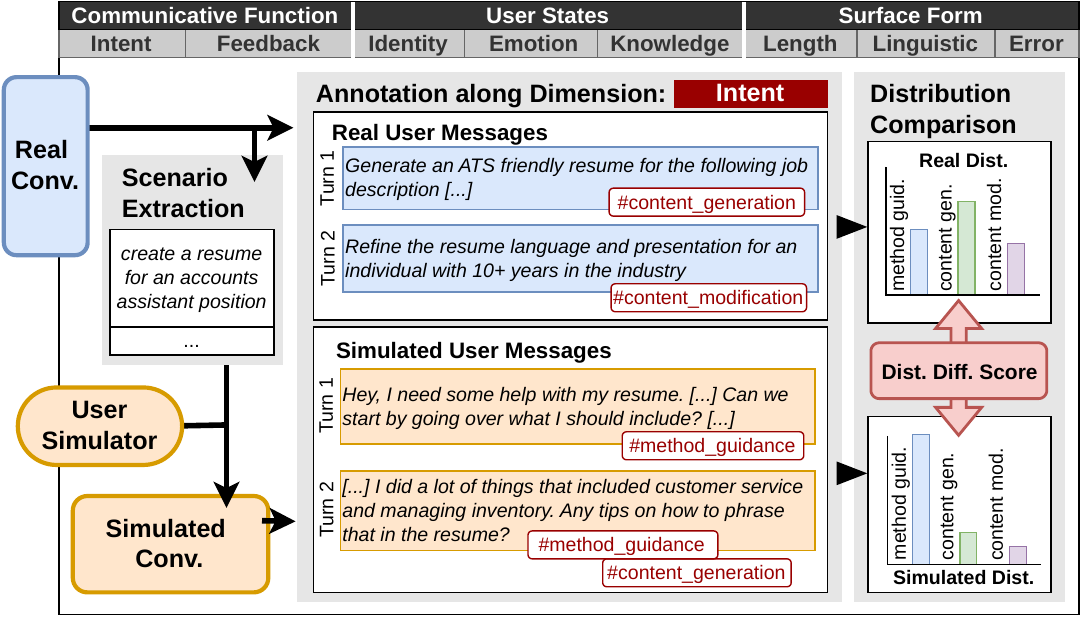}
    \caption{Illustration of the \texttt{realsim} framework.}
    \label{fig: main_diagram}
\end{figure*}

As illustrated in Figure~\ref{fig: main_diagram}, \FrameworkName\space requires simulated users to interact with the assistant chatbot in the same task scenarios as real users did. Simulated and real dialogues are all annotated along the 8 dimensions, and the distributions of annotation results are finally compared. Through our experiments, we find that simulated users seem to be easier interlocutors: they tend to introduce less communication friction, provide much more positive feedback and disclose more contextual information. Evaluation methods that rely on user simulation may thus produce optimistic assessment of chatbot capabilities. We also find that simulated users struggle to capture some domain-specific real user behaviors, highlighting the potential need to design domain-specific user simulators. 

\section{Related Work}
\subsection{User Simulation for (AI) Chatbot Evaluation}

To evaluate the performance of chatbots in multi-turn interactions, prior work explored user simulations to elicit chatbot behaviors. 
Some simulation approaches rely on \textbf{prompting LLMs} by providing user and/or task-related information as model instructions. In $\tau$-Bench \citep{yao2024taubenchbenchmarktoolagentuserinteraction}, for example, a LLM was prompted to simulate an airline or retail customer, in order to evaluate chatbot ability to follow domain-specific rules such as company policies. Simulating different user types is explored in MINT \citep{MINT} and in IQA-Eval \citep{IQA-EVAL}, where a LLM was prompted to act like an ``informative user,'' or an ``expert.'' More complex simulation instructions are involved in the work of \citet{svikhnushina-pu-2023-approximating}, where LLMs are instructed to take the role of a person feeling a specific emotion and experiencing a specific situation, in order to evaluate social chatbots. 

Apart from prompting, some prior work explored \textbf{specifically training LLMs} to act like human users. For example, ChatBench \citep{ChatBench} evaluates the collaboration between real users and LLMs in problem-solving by fine-tuning a LLM to simulate users' i) initial query, ii) follow-up messages, and iii) decisions to end the conversation with the answer to the question being solved. There are also prior work focused on building user simulators for chatbot evaluation or for informing the design of AI features, such as UserLM \citep{naous2025flippingdialoguetrainingevaluating} and HumanLM \citep{wu2026humanlmsimulatingusersstate}.

Considering that recent work have identified flaws and potential harms arising from the use of LLM-based user simulation in social science studies \citep{Wang2025-at, not_yet}, our field needs to also be cautious about its use in chatbot evaluation. Our proposed framework aims to provide practitioners with more insights on the behaviors of user simulators, so they can make more informed judgment about their usage in chatbot evaluation. 

\subsection{Evaluation of Simulated Users}
The evaluation of user simulation within the context of chatbot evaluation is limited, as the quality of evaluation methods that involves user simulation is often only assessed in terms of \textit{concurrent validity}, where a benchmark involving user simulation is considered good if 
its evaluation outcomes are similar to outcomes obtained through the use of real dialogues \citep{li2019acuteevalimproveddialogueevaluation, IQA-EVAL}. The validation of these methods thus often misses \textit{content validity}: assessing whether simulated dialogues in these methods actually reflect real user-chatbot interactions. Even when such validation is done, the current methodology is very limited: in ChatBench \citep{ChatBench}, for example, the simulator's first utterance is compared against the real user's first utterance lexical overlap metrics. This only provides a coarse quality signal of the first turn, not taking into account the subsequent interaction.  

Existing frameworks for evaluating synthetic dialogues involve labelling each turn of the dialogue and analyzing label frequencies. The framework of \citet{ivey2024realroboticassessingllms} contains correlation-based measures, applied on e.g., prompt classification labels and sentiment classification labels. DialogFlowPPL metric from SDialog \citep{burdisso2026sdialogpythontoolkitendtoend} constructs a dialogue state flow graph from a set of real dialogues and then computes a perplexity-like score for each synthetic dialogue to measure how it fits the graph. Our framework focus on computing and analyzing distribution differences along fine-grained dimensions that influence interaction experience, including but not limited to user intent type, feedback type, and domain-specific knowledge. 

\section{\FrameworkName\space Framework}
\label{sec: framework}
Our framework proposes to evaluate the realism of a user simulator by identifying and comparing patterns in its interaction against the behavior patterns observed in real user-chatbot dialogues. We thus take a distributional view, rather than evaluating the realism of synthetic dialogues at the sample-level. For a given set of real dialogues, we first use the user simulator under evaluation to create a set of simulated dialogues in parallel (Section~\ref{subsec: scenario_simulation}). Then, we annotate each dialogue (real or simulated) along 8 dimensions of realism: \textit{Length, Intent, Feedback, Emotion, Linguistics, Identity, Knowledge, and Error} (Section~\ref{subsec: realism_dim}). Finally, we qualitatively and quantitatively compare the annotation results of simulated vs. real dialogues through descriptive statistics, distribution difference, semantic similarity, and across-domain correlation scores (Section~\ref{subsec: comparison}). This process is illustrated in Figure~\ref{fig: main_diagram}. 

\subsection{Scenario-Based Simulation}
\label{subsec: scenario_simulation}
Given a set of real user-chatbot dialogues, we first extract a scenario description from each dialogue by summarizing the main task that the user is trying to accomplish (e.g., ``create a resume for an accounts assistant position’’). We then produce simulated dialogues by instructing the user simulator to interact with an assistant chatbot while aiming to accomplish the same task (e.g., ``simulate a user interacting with a chatbot to create a resume...’’). This process aims to ensure that the sets of simulated and real dialogues are comparable to one another in the same scenario. 

\subsubsection{Dataset of Real Interactions}
\label{sec: dataset_descrip}
Our framework includes a dataset of 1K real user-chatbot dialogues curated from WildChat \citep{zhao2024wildchat1mchatgptinteraction} and LMSYS-1M \citep{zheng2024lmsyschat1mlargescalerealworldllm}, where we selected examples where users take 4 or more turns to accomplish a single, coherent task with the chatbot (rather than a sequence of unrelated queries). Given that user behaviors likely differ across different domains, we also ensure a wide domain coverage. Our dataset covers 16 domains, including \textbf{informatics-related} domains (computer hardware, cybersecurity, data storage and analysis methods), \textbf{health domains} (general health, mental health and nutrition, extracted from HealthChat \citep{paruchuri-etal-2025-whats}), \textbf{creative domains} (brainstorming of brand names, song and poem writing, discussion or recommendation about books, movies, shows, or music), and \textbf{other} tasks related to work or personal life (help with job application, emails, travel planning, finance-related information or advice, usage related to education and learning, discussion about law, history and politics). We describe the data curation and data distribution in more details in Appendix~\ref{app: data_curation}. 

This dataset allows us to evaluate the performance of user simulators across multiple domains. Practitioners using our framework could choose to use other datasets of real dialogues that are more representative of their chatbots' use context, as long as they are also multi-turn and task-focused.

\subsection{Dimensions of Realism}
\label{subsec: realism_dim}
The user simulation literature outlines three aspects of user behavior that realistic simulation must capture: the communicative actions users perform, the characteristics users bring to the interaction, and how those actions are linguistically expressed~\cite{schatzmann2006survey, zukerman2001natural}. 
Each aspect distinctly evaluates chatbot performance: dialogue strategy, user adaptiveness, and language understanding, respectively.

We ground our decomposition of realism into 8 dimensions along these three established levels.
At the level of \textbf{communicative functions}, we capture the sub-goals users pursue (\textit{Intent}) and the evaluative reactions they provide to chatbot responses (\textit{Feedback}). 
At the level of \textbf{user states}, we capture the attributes users bring to or reveal during interaction: the emotions they express (\textit{Emotion}), the personal context they disclose (\textit{Identity}), and the domain knowledge they possess (\textit{Knowledge}). 
At the level of \textbf{surface form}, we assess how user messages are realized as text: the amount of text users produce per interaction (\textit{Length}), their readability and lexical complexity (\textit{Linguistics}), and the errors and disfluencies that naturally arise in human writing (\textit{Error}). These dimensions are further motivated by human-computer interaction research on chatbots, which has shown that user-side factors spanning all three levels, such as feedback patterns, emotional expression, self-disclosure, and communication style, significantly influence conversation outcomes and experiences \citep{folstad2020users,chaves2021should,xiao2020if,sharma2026feedback}.

\subsubsection{Communicative Functions}
\para{Intent} The main task often gets decomposed into sub-goals. For example, when a user plans a trip, they may first ask the chatbot general questions about their destination before requiring it to generate a detailed itinerary. These sub-goals may also emerge in reaction to chatbot responses, e.g., asking for definition of unfamiliar terms used by the chatbot. Given that we need to evaluate the different capabilities involved in accomplishing these sub-goals, it is important that simulated interactions mirror the variation of real user intent. For this dimension, we label each user turn using the taxonomy of \citet{shelby2025taxonomyuserneedsactions}, which categorizes instrumental user goals into: \textit{information retrieval, information discovery, information clarification, information distillation, information analysis, procedural guidance, content generation, and content modification}; one turn can have more than one Intent tags.

\para{Feedback}
In multi-turn settings, it is important to evaluate how chatbots process various user feedback, such as explicit negative feedback (e.g., ``that was too long'') or regeneration requests (e.g., ``try again''). Simulated interactions thus need to reflect the various types of feedback that chatbots may encounter in real deployment settings. For this dimension, we label each user turn with the following categories, inspired from the taxonomy of \citet{shelby2025taxonomyuserneedsactions}: \textit{no feedback}, \textit{explicit negative feedback}, \textit{explicit positive feedback}, \textit{regeneration request} which can be interpreted as implicit negative feedback, \textit{continuation request} which we interpret could signal chatbot response incompleteness, \textit{requesting clarification} and \textit{providing clarification} which could signal dialogue breakdown. 

\subsubsection{User States}
\para{Emotion} Users may express various emotions in their messages, and how chatbots respond to them is especially important to evaluate in domains such as customer service and mental health. Simulated users thus need to mirror the emotional expression (or lack thereof) of real users. For this dimension, we use an emotion classification model to categorize each sentence of user messages into \textit{anger}, \textit{disgust}, \textit{fear}, \textit{joy}, \textit{sadness}, \textit{surprise}, or \textit{neutral}\footnote{\url{https://huggingface.co/j-hartmann/emotion-english-distilroberta-base}; The label of \textit{neutral} is also used when the prediction confidence is below 75\%.}.

\para{Identity}
When interacting with chatbots, people sometimes reveal information about themselves, e.g., their demographic information and physical conditions when asking the chatbot for general health advice. How chatbot process and elicit such information is an important capability to assess. Simulated users should thus mirror real user behavior by providing information that could plausibly belong to a real human, simulating a sense of personal ``identity.'' For this dimension, we focus on whether the user messages contain explicit mentions of: \textit{demographic information} such as age, gender, or occupation; \textit{physical information} which includes characteristics such as height, weight, allergies, or diseases; \textit{interpersonal relationships} such as family, friends, employers or clients.; \textit{past} events or activities; \textit{future} plans, events, or activities; \textit{worldview} which covers the user's beliefs such as political views. 

\para{Knowledge}
User knowledge influences how they interact with chatbots, especially for tasks involving information retrieval and processing. For example, users with different level of domain expertise may ask very different questions for the same health-related topic. To evaluate whether chatbots can handle varying level of user knowledge, user simulators need to reflect the variety of user knowledge found in real user interactions. For this dimension, we 
prompt an LLM to list statements about what the user does and does not seem to know for a given dialogue. For example, if the users asked ``how to manage diabetes,'' we can infer that the user knows about diabetes and it being a disease needing daily management, but that they do not know the steps involved in managing it. We compare the semantic similarity of statements extracted from simulated dialogues to those of real dialogues.  

\subsubsection{Surface Form}
\para{Length} Real users can only communicate a limited volume of information, as the effort of communication costs time and energy. To enable a better evaluation of whether (or to what extent) the chatbot can achieve task success under this constraint, user simulation needs to mirror real user message length. For this dimension, we compute for each dialogue the total word count of user messages and the number of user turns. 

\para{Linguistic} LLMs often adopt language patterns that distinguish their output from real human-written text. To evaluate whether chatbots can effectively communicate with people, it is important that LLM-based user simulators are able to adopt human-like linguistic patterns. We are thus interested in measuring the linguistic characteristics of real vs. simulated interactions. For this dimension, we compute the Flesch–Kincaid readability score \citep{KincaidJP1975DoNR} as well as the Measure of Textual Lexical Diversity score \citep{MTLD_McCarthy2010-ht}, with all user messages being considered as one corpus. 

\para{Error} Error introduces friction to interactions, leading to misunderstandings and communication breakdowns. To ensure user satisfaction, it is important for chatbots to be able to resolve or repair these potential communication failures. As a result, user simulators should also be able to replicate human errors such as grammatical mistakes. For this dimension, we use a rule-based grammar checker\footnote{\url{https://language-tool-python.readthedocs.io/en/latest/}} to detect and count potential errors in user messages. 

\subsection{Comparison Methods}
\label{subsec: comparison}
After annotating real and simulated dialogues along the 8 dimensions, we compare the annotation results mainly through i) descriptive statistics, ii) distribution difference scores, and iii) across-domain correlation scores. 

Descriptive statistics allow for a qualitative judgment of distribution similarity, while distribution difference scores quantify it. For Intent, Feedback, Emotion, and Identity, we aggregate annotation labels into categorical distributions, which are compared against one another through \textbf{total variation distance} scores. For Length and Error, their annotation results are discrete distributions that are compared against one another through 
\textbf{Wasserstein distance} scores. For Knowledge, given that the annotation results are texts, we measure the semantic similarity between statements extracted from simulated conversations vs. real conversations, for each domain in our dataset.\footnote{Semantic similarity computed through cosine similarity of \texttt{all-MiniLM-L6-v2} embeddings.}

Given that our dataset of real dialogues span many domains (Section~\ref{sec: dataset_descrip}), we are also able to make domain-specific distribution comparisons (e.g., X\% of real dialogues vs. Y\% of simulated dialogues in the domain of mental health reveal \textit{interpersonal relationships}). Moreover, we can assess how realistically the behaviors of user simulators vary across different domains via \textbf{across-domain correlation scores}. We first compute domain-specific values for each dimension (e.g., frequency of Identity = \textit{interpersonal relationships} in mental health domain) and then compute the Pearson correlation coefficient between values obtained from simulated conversations vs. real conversations.

\section{Experiments}
\label{sec: experimental_setup}
To illustrate the use of the our framework and showcase insights that one could obtain from it, we evaluates 7 different user simulation methods, covering fine-tuning and persona-based prompting strategies: i) \textbf{UserLM} \citep{naous2025flippingdialoguetrainingevaluating}, a LLM trained on WildChat\citep{zhao2024wildchat1mchatgptinteraction}\footnote{Given that our curated dataset also contains dialogues from the WildChat dataset, we expect UserLM to have greater performance.} to produce user utterances; ii) GPT-4o-mini, GPT-5.2 and Gemma 3 seeded with the \textbf{generic persona} of ``a human user''; iii)  GPT-4o-mini, GPT-5.2 and Gemma 3 seeded with an \textbf{informed persona} which is generated to be relevant to each specific scenario (e.g., ``a newly diagnosed adult with type 2 diabetes'' for a scenario about diabetes management). The prompt templates for persona generation and the implementation details of user simulators are presented in Appendix~\ref{app: user_simulation}. 

The assistant chatbot that all simulators interact with is a ``vanilla'' GPT-4o-mini, without any system instructions. For dimensions needing LLM annotation (i.e., Intent, Feedback, Identity, and Knowledge) we prompt GPT-5 using the prompt templates provided in Appendix~\ref{app: LLM_annotation}, where we also describe our process in validating our methodology through human annotation on a development set.

\section{Results}
\subsection{General Insights}
\begin{figure*}[ht!]
    \centering
    \includegraphics[scale = 0.3]{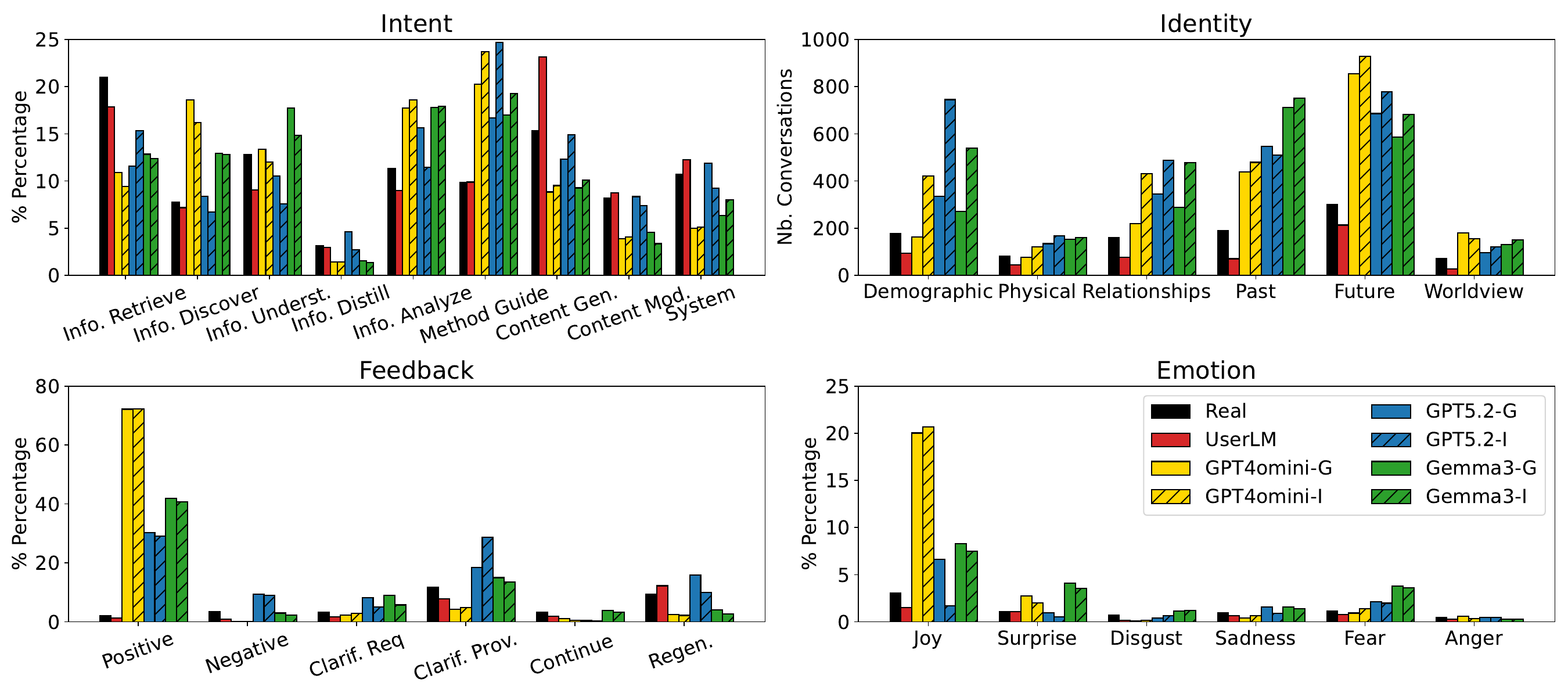}
    \caption{Category Frequency for Intent, Identity Feedback, and Emotion.}
    \label{fig: categorical_bargraph}
\end{figure*}
\begin{table}[t]
\centering
\small
\setlength{\tabcolsep}{5pt}
\begin{tabular}{lccccc}
& \multicolumn{2}{c}{\textbf{Length}} & \multicolumn{2}{c}{\textbf{Linguistic}} & \textbf{Error}\\
&\textbf{Avg. \#Turn} & \textbf{Avg. \#Word/Conv} & \textbf{Readability} &\textbf{ Lexical Rich.} & \textbf{Rate} \\
Real  &	7.01	&	150 &	11.36	&	59.0	&	9.17\\
UserLM &	7.43	&	90	&	10.10	&	21.4 &	10.77\\
GPT4omini-G &	7.01	&	418 &	6.82	&	112.4	&	2.06\\
GPT4omini-I &	6.34	&	524	&	7.71	&	117.4 &	2.95\\
GPT5.2-G &	5.03	&	659	&	9.29	&	153.4 &	9.91\\
GPT5.2-I &	7.01	&	2102	&	12.14	&	129.5	&	6.52\\
Gemma3-G &	4.32	&	309	&	6.38	&	112.6 &	2.00\\
Gemma3-I &	5.23	&	412	&	7.38	&	119.5	&	1.98 \\

\end{tabular}
\caption{Descriptive Statistics for Length, Linguistic, and Error.}
\label{table: descriptive_stats}
\end{table}
\textbf{\textit{Simulated users tend to be easier for chatbots to interact with, which may result in an over-optimistic assessment of chatbot capabilities.}} In terms of their \textit{surface form}, simulated user messages tend to be more wordy, more readable, with fewer grammatical errors than real user messages (Table~\ref{table: descriptive_stats}). As for the \textit{user states}, we observe from Figure~\ref{fig: categorical_bargraph} that simulated users tend to disclose more information, often revealing their pretended relationships with others, past actions and future plans. They are also more emotionally expressive, often expressing joy. 
These observations suggest that simulated users may be less challenging for chatbots to interact with, as they struggle to introduce communication frictions, such as the lack of contextual information. Chatbot evaluation methods that rely on simulated user messages might thus overestimate chatbot performance and overlook their limitations.

\textbf{\textit{Simulated users decompose tasks differently and are more easily satisfied than real users, threatening the validity of chatbot evaluation methods that rely on them.}} In terms of \textit{communicative functions}, we observe that real user intents fall most frequently under information retrieval and content generation, while simulated users frequently seek method guidance and information analysis. We also observe discrepancies in the Intent category of information discovery, which appear much more frequently in messages produced by GPT4o-mini and Gemma3 methods than in real user messages. As a result of these discrepancies, chatbot evaluation methods that rely on simulated users may thus target different chatbot capabilities than what is most necessary for successful interactions with real users. 
Furthermore, simulated users tend to contain much more positive feedback to the chatbot, making it difficult to observe and evaluate how chatbots respond to negative user feedback, or how they manage the lack of explicit user feedback. 

\textbf{\textit{Fine-tuning on real user messages may help replicate real user behavior trends.}} Through distribution difference scores (Table~\ref{table: distribution_diff}), we can quantitatively compare between different user simulators. UserLM, which is trained on real user messages, is generally able to more closely reflect real user behavior patterns. There is however room for improvement when it comes to turn length distribution, i.e., realistically deciding when to end conversations. 

\begin{table}[!ht]
\centering
\small
\setlength{\tabcolsep}{5pt}
\begin{tabular}{l|cccc|ccc}
& \multicolumn{4}{c|}{\textbf{Categorical Dist. Diff.}}&  \multicolumn{3}{c}{\textbf{Discrete Dist. Diff.}}\\
& \textbf{Intent} & \textbf{Feedback} & \textbf{Emotion} & \textbf{Identity} & \textbf{\#Turn} & \textbf{\#Word} &  \textbf{\#Error} \\
UserLM		 &\textbf{9.9}	 & \textbf{6.7}	 &3.0 &\textbf{22.7}             & 1.52 & \textbf{60.4} &\textbf{1.87}\\
GPT4omini-G		 &28.3		 &45.4		 &18.8	 &49.6           &0.91	 & 267.8 &7.11 \\
GPT4omini-I		 &29.6	 &45.3 &18.7	 &77.7          &\textbf{0.70}    &373.3 &6.22  \\
GPT5.2-G		 &14.7	 &27.5	 &5.2	 &58.0            &1.98 &509.6 &4.02  \\
GPT5.2-I		 &14.9		 &27.3		 &\textbf{2.1}	 &91.5            &1.23  &1951.3	 &3.05\\
Gemma3-G		 &23.8		 &27.6		 &11.9		 &57.9                 &2.69	 &208.8 &7.33 \\
Gemma3-I		 &23.2		 &25.4	 &10.3	 &89.0                    &1.78	 &287.5 &7.33 \\
\end{tabular}

\caption{Distribution Difference Scores. Total Variation Distance score is used for categorical distributions, while Wasserstein-1 Distance score is used for discrete distributions.}
\label{table: distribution_diff}
\end{table}

\subsection{Domain-Specific Insights}
\label{sec: results_domain_specific}
\begin{figure*}[h!]
    \centering
    \includegraphics[scale = 0.39]{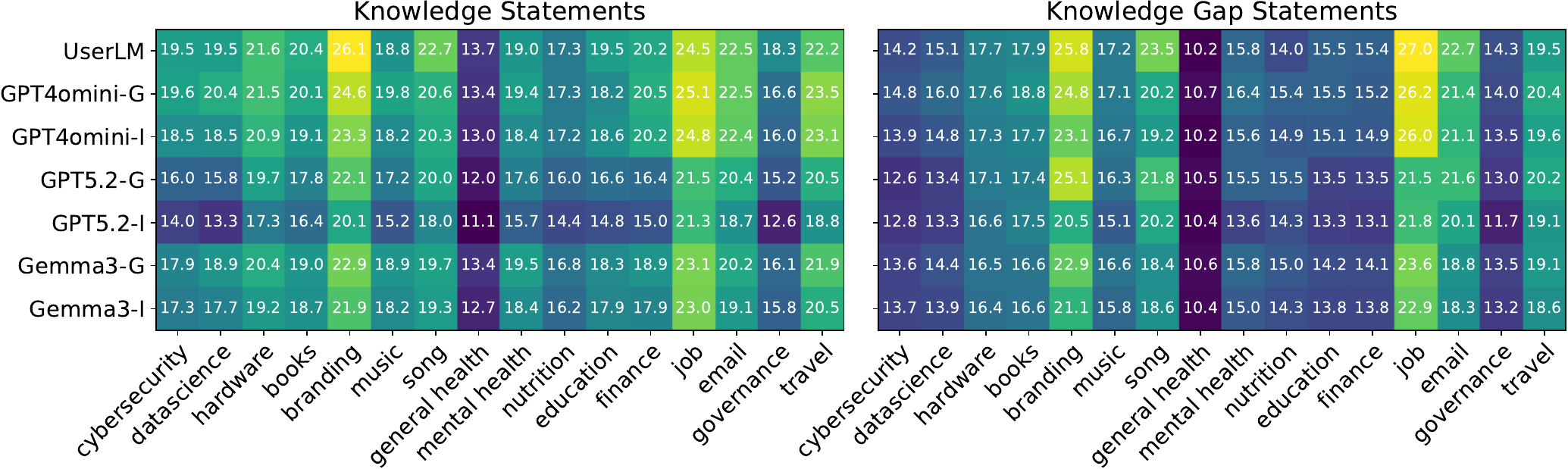} 
    \caption{Heatmap of Knowledge Statements Semantic Similarity Scores Between Real vs. Simulated Conversations.}
    \label{fig: heatmap}
\end{figure*}
\begin{figure*}[h!]
    \centering
    \includegraphics[scale = 0.33]{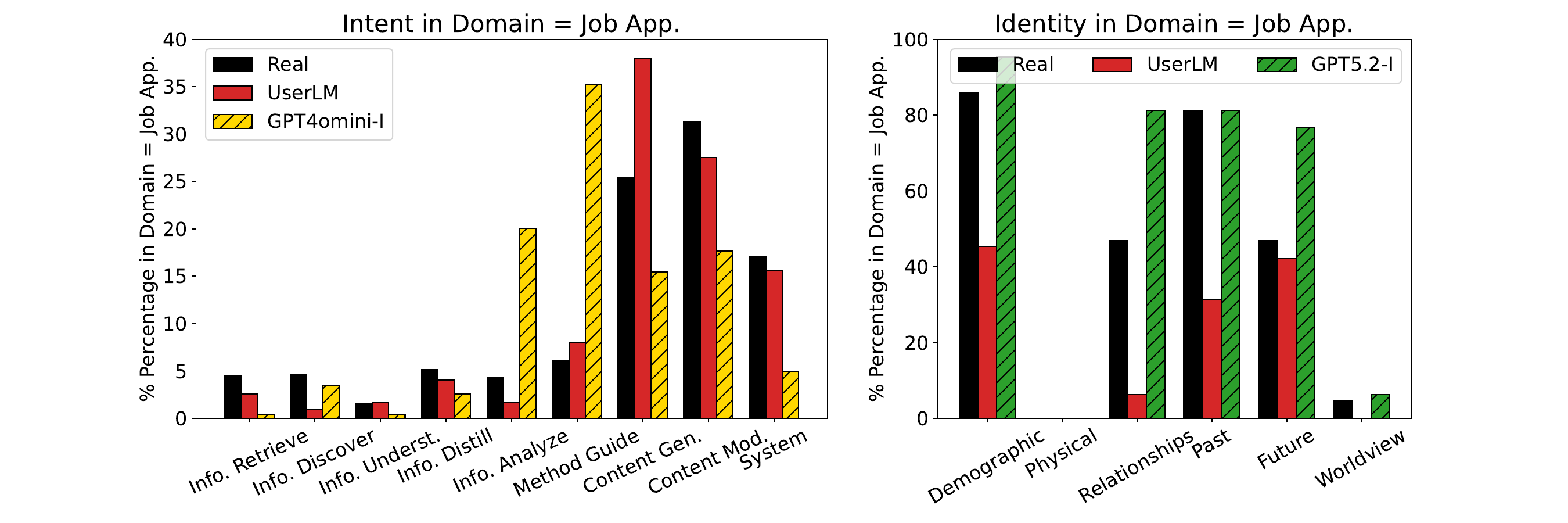} 
    \caption{Intent and Identity Distributions for Job Application Domain.}
    \label{fig: heatmap}
\end{figure*}

\textbf{\textit{Some domains tend to be more challenging to simulate, perhaps requiring more tailored design of user simulation methods.}}
Through the heatmap of the semantic similarity scores between Knowledge statements extracted from real vs. simulated user messages (Figure~\ref{fig: heatmap}), we can observe that simulated users seem to better match real user Knowledge in the domains of branding brainstorming and job application, while having worse performance in conversations about general health and governance discussion topics. Through qualitative examples in the domain of general health (Appendix~\ref{app: examples}), we observe that Knowledge statements from real messages seem to be richer and more specific, suggesting that user simulators might need to be explicitly provided with more contextual information about the task.

\textbf{\textit{User simulators can struggle to capture some domain-specific real user behaviors.}}
Our framework supports domain-level analysis for dimensions other than Knowledge. For example, for the domain of job application, we can observe that GPT4o-mini performs poorly along the dimension of Intent: requesting for method guidance (e.g., how to improve a resume), while real conversations focus on content generation and modification (e.g., generate a resume). In the same domain, UserLM has poor performance along the dimension of Identity: it failed to reveal information that real users do when working on their job application, such as interpersonal relationships (e.g., past clients, coworkers). To ensure that user simulators capture such kinds of domain-specific real behaviors, domain-specific evaluation of user simulation may be necessary and especially relevant if the user simulator is intended to help evaluate a chatbot developed for a specific domain of application\footnote{We selected the domain of job application as an example here because, through visualizing the Intent and Identity distribution difference scores for each domain (Appendix~\ref{app: examples}), we identified that GPT4o-mini simulators had the worst performance in the domain of job application along the Intent dimension, while UserLM performed poorly in the same domain along the Identity dimension.}.

\subsection{Across-Domain Variation}

\begin{figure*}[t!]
    \centering
    \includegraphics[scale = 0.34]{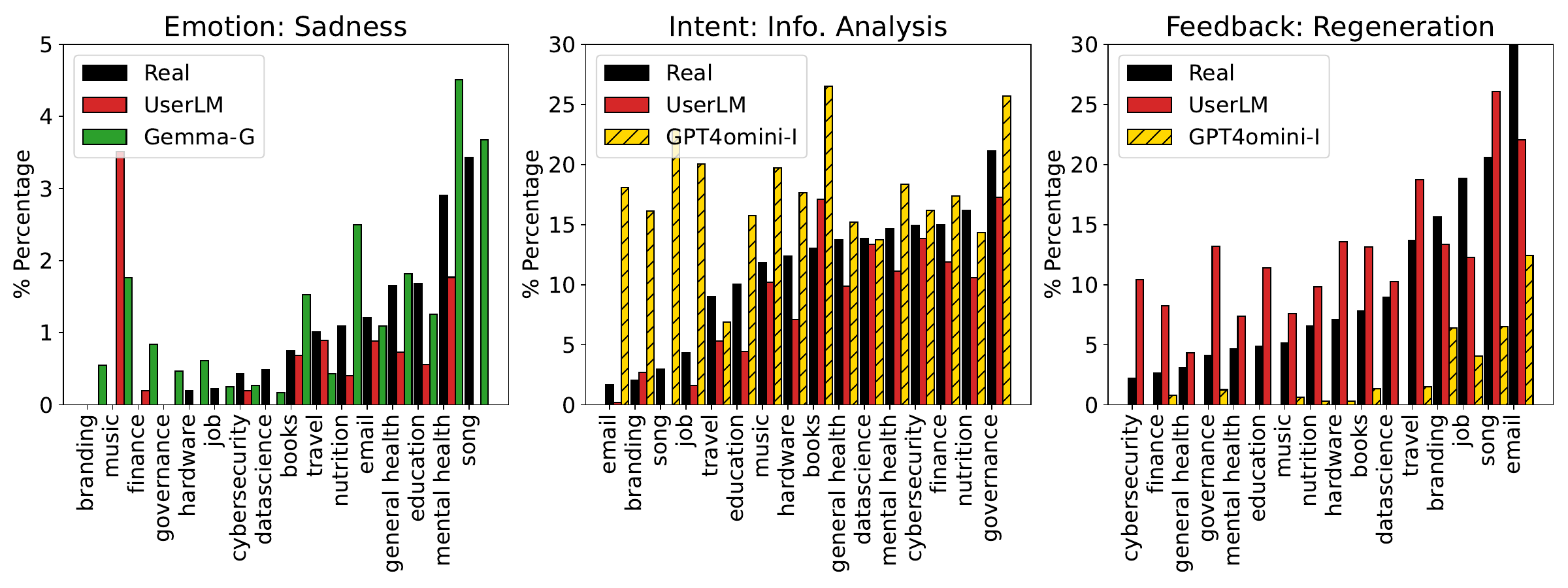} 
    \caption{Frequency Variation Across Domains (domains ordered according to Real frequency values)}
    \label{fig:example_corr}
\end{figure*}

\textbf{\textit{Real user behaviors can vary greatly across domains, and ``general-purpose'' user simulators would need to capture this variation.}} As shown in Figure~\ref{fig:example_corr}, real users express sadness more often when writing songs and poems, than when they do when brainstorming for brand names, or looking for music recommendations. UserLM, which seems to be the overall best user simulator through our analysis thus far, fails at capturing this variation. Together with our previous observations, this illustrates the challenge of building a user simulator that is applicable across different domains.

\begin{table}[!t]
\centering
\small
\setlength{\tabcolsep}{5pt}
\begin{tabular}{l|cccc|ccc}
& \multicolumn{4}{c|}{\textbf{Correlation Score Averaged }}&  \multicolumn{3}{c}{\textbf{Correlation Score}}\\
& \textbf{Intent} & \textbf{Feedback} & \textbf{Emotion} & \textbf{Identity} & \textbf{\#Turn} & \textbf{\#Word} &  \textbf{\#Error} \\
UserLM	&	\textbf{91.7} &	\textbf{58.6}&	46.7	&	\textbf{88.6}	&	\textbf{65.8}	&	\textbf{41.8}&	58.1 \\
GPT4omini-generic	&	82.5	&	45.4	&	50.4	&	71.6	&	18.1	&	-1.5	&	65.2\\
GPT4omini-informed	&	77.2	&	30.7	&	52.2	&	61.0	&	12.9	&	4.4	&	42.8\\
GPT5.2-generic	&	87.9	&	36.1	&	68.3	&	63.7	&	9.7	&	-1.3	&	\textbf{72.8}\\
GPT5.2-informed	&	79.8	&	28.8	&	\textbf{72.5}&	52.2	&	-27.8	&	22.4	&	53.8\\
Gemma3-generic	&	87.7	&	34.4	&	53.7	&	75.1	&	33.4	&	11.4	&	71.7\\
Gemma3-informed	&	84.4	&	33.9	&	54.8	&	59.8	&	15.4	&	26.2 &	71.2\\
\end{tabular}

\caption{Across-Domain Correlation Results. }
\label{table: distribution_statistical_test}
\end{table}

\section{Conclusion}
For user simulation to be a useful methodology in chatbot evaluation, simulated users need to capture real user behaviors when interacting with chatbots. We propose \FrameworkName\space to help practitioners assess the extent to which simulated user behavior patterns mirror real ones. Our framework cover three main aspects of user behavior: communicative functions, user states, and surface form of user messages. We find that existing user simulation methods, especially those relying on prompting LLMs, lack realism and thus cannot adequately support chatbot evaluation. 

\section*{Reproducibility Statement}
To support the reproducibility of our work, we describe in detail our data curation, user simulation, and LLM-based annotation methodologies in Appendix~\ref{app: data_curation}, \ref{app: user_simulation}, and \ref{app: LLM_annotation}, where we also include all the prompt templates that we used in our experiments. The dataset of real dialogues, the set of extracted scenario descriptions, the set of generated persona informed by the scenario descriptions, and the set of simulated dialogues are all published at \url{https://github.com/isle-dev/realsim}.

\section*{Ethics Statement}
\para{Limitations} Our findings are limited by the data and by the user simulation methods that we have included in our experiments. More specifically, considering that our dataset is curated from WildChat \citep{zhao2024wildchat1mchatgptinteraction} and LMSYS-1M \citep{zheng2024lmsyschat1mlargescalerealworldllm}, it might not be representative of how people currently use chatbots. However, our framework can be applied to other human-AI conversation datasets. If practitioners have access to data of real dialogues that are more recent and representative of their specific domain of chatbot application, we encourage them to use their data to instantiate our framework. Moreover, our framework can be further adapted to specific domains by operationalizing the 8 dimensions of \FrameworkName\space in a more domain-specific manner. For example, in the domain of healthcare, Intent categories can be defined to more precisely align with users' medical needs. 

\para{Simulating humans in research}
Simulating human behaviors instead of involving real human participants can be harmful in various ways. By failing to effectively capture human behaviors and attitudes, simulated data may lead studies that rely on them to produce invalid conclusions \citep{not_yet}. Moreover, simulation can lead to representational harms, where marginalized demographic groups are misrepresented by simulated data intended to represent their behaviors or attitudes. For example, simulation may essentialize marginalized identities, (re)producing caricatures \citep{Wang2025-at, cheng-etal-2023-compost}. Our framework could help practitioners identify whether user simulation effectively capture the behaviors of real human users interacting with chatbots and make a more informed decision about whether user simulation should be used in their research or development of chatbots, and if so, how much they can rely on the evaluation outcomes.

\para{Human-like LLMs} 
Outside of research and development, LLMs that closely imitate human behaviors can be misused, such as facilitating impersonation scams. Our work focuses on user simulation, where LLMs aim to imitate how people interact with chatbots, as opposed to how people interact with other people. Despite this distinction, our work may be misused to develop more human-like LLMs for harmful purposes. 

\bibliography{colm2026_conference}

@misc{shelby2025taxonomyuserneedsactions,
      title={Taxonomy of User Needs and Actions}, 
      author={Renee Shelby and Fernando Diaz and Vinodkumar Prabhakaran},
      year={2025},
      eprint={2510.06124},
      archivePrefix={arXiv},
      primaryClass={cs.HC},
      url={https://arxiv.org/abs/2510.06124}, 
}

@inproceedings{paruchuri-etal-2025-whats,
    title = "``What{'}s Up, Doc?'': Analyzing How Users Seek Health Information in Large-Scale Conversational {AI} Datasets",
    author = "Paruchuri, Akshay  and
      Aziz, Maryam  and
      Vartak, Rohit  and
      Ali, Ayman  and
      Uchehara, Best  and
      Liu, Xin  and
      Chatterjee, Ishan  and
      Agrawal, Monica",
    editor = "Christodoulopoulos, Christos  and
      Chakraborty, Tanmoy  and
      Rose, Carolyn  and
      Peng, Violet",
    booktitle = "Findings of the Association for Computational Linguistics: EMNLP 2025",
    month = nov,
    year = "2025",
    address = "Suzhou, China",
    publisher = "Association for Computational Linguistics",
    url = "https://aclanthology.org/2025.findings-emnlp.125/",
    doi = "10.18653/v1/2025.findings-emnlp.125",
    pages = "2312--2336",
    ISBN = "979-8-89176-335-7"
}

@misc{naous2025flippingdialoguetrainingevaluating,
      title={Flipping the Dialogue: Training and Evaluating User Language Models}, 
      author={Tarek Naous and Philippe Laban and Wei Xu and Jennifer Neville},
      year={2025},
      eprint={2510.06552},
      archivePrefix={arXiv},
      primaryClass={cs.CL},
      url={https://arxiv.org/abs/2510.06552}, 
}

@misc{zhao2024wildchat1mchatgptinteraction,
      title={WildChat: 1M ChatGPT Interaction Logs in the Wild}, 
      author={Wenting Zhao and Xiang Ren and Jack Hessel and Claire Cardie and Yejin Choi and Yuntian Deng},
      year={2024},
      eprint={2405.01470},
      archivePrefix={arXiv},
      primaryClass={cs.CL},
      url={https://arxiv.org/abs/2405.01470}, 
}

@misc{zheng2024lmsyschat1mlargescalerealworldllm,
      title={LMSYS-Chat-1M: A Large-Scale Real-World LLM Conversation Dataset}, 
      author={Lianmin Zheng and Wei-Lin Chiang and Ying Sheng and Tianle Li and Siyuan Zhuang and Zhanghao Wu and Yonghao Zhuang and Zhuohan Li and Zi Lin and Eric P. Xing and Joseph E. Gonzalez and Ion Stoica and Hao Zhang},
      year={2024},
      eprint={2309.11998},
      archivePrefix={arXiv},
      primaryClass={cs.CL},
      url={https://arxiv.org/abs/2309.11998}, 
}

@inproceedings{cheng-etal-2023-compost,
    title = "{C}o{MP}os{T}: Characterizing and Evaluating Caricature in {LLM} Simulations",
    author = "Cheng, Myra  and
      Piccardi, Tiziano  and
      Yang, Diyi",
    editor = "Bouamor, Houda  and
      Pino, Juan  and
      Bali, Kalika",
    booktitle = "Proceedings of the 2023 Conference on Empirical Methods in Natural Language Processing",
    month = dec,
    year = "2023",
    address = "Singapore",
    publisher = "Association for Computational Linguistics",
    url = "https://aclanthology.org/2023.emnlp-main.669/",
    doi = "10.18653/v1/2023.emnlp-main.669",
    pages = "10853--10875"
}

@misc{ivey2024realroboticassessingllms,
      title={Real or Robotic? Assessing Whether LLMs Accurately Simulate Qualities of Human Responses in Dialogue}, 
      author={Jonathan Ivey and Shivani Kumar and Jiayu Liu and Hua Shen and Sushrita Rakshit and Rohan Raju and Haotian Zhang and Aparna Ananthasubramaniam and Junghwan Kim and Bowen Yi and Dustin Wright and Abraham Israeli and Anders Giovanni Møller and Lechen Zhang and David Jurgens},
      year={2024},
      eprint={2409.08330},
      archivePrefix={arXiv},
      primaryClass={cs.CL},
      url={https://arxiv.org/abs/2409.08330}, 
}

@misc{li2019acuteevalimproveddialogueevaluation,
      title={ACUTE-EVAL: Improved Dialogue Evaluation with Optimized Questions and Multi-turn Comparisons}, 
      author={Margaret Li and Jason Weston and Stephen Roller},
      year={2019},
      eprint={1909.03087},
      archivePrefix={arXiv},
      primaryClass={cs.CL},
      url={https://arxiv.org/abs/1909.03087}, 
}

@inproceedings{MINT,
 author = {Wang, Xingyao and Wang, Zihan and Liu, Jiateng and Chen, Yangyi and Yuan, Lifan and Peng, Hao and Ji, Heng},
 booktitle = {International Conference on Learning Representations},
 editor = {B. Kim and Y. Yue and S. Chaudhuri and K. Fragkiadaki and M. Khan and Y. Sun},
 pages = {32593--32627},
 title = {MINT: Evaluating LLMs in Multi-turn Interaction with Tools and Language Feedback},
 url = {https://proceedings.iclr.cc/paper_files/paper/2024/file/8a0d3ae989a382ce6e50312bc35bf7e1-Paper-Conference.pdf},
 volume = {2024},
 year = {2024}
}

@inproceedings{ChatBench,
   title={ChatBench: From Static Benchmarks to Human-AI Evaluation},
   url={http://dx.doi.org/10.18653/v1/2025.acl-long.1262},
   DOI={10.18653/v1/2025.acl-long.1262},
   booktitle={Proceedings of the 63rd Annual Meeting of the Association for Computational Linguistics (Volume 1: Long Papers)},
   publisher={Association for Computational Linguistics},
   author={Chang, Serina and Anderson, Ashton and Hofman, Jake M.},
   year={2025},
   pages={26009–26038} }

@inproceedings{IQA-EVAL,
author = {Li, Ruosen and Li, Ruochen and Wang, Barry and Du, Xinya},
title = {IQA-EVAL: automatic evaluation of human-model interactive question answering},
year = {2024},
isbn = {9798331314385},
publisher = {Curran Associates Inc.},
address = {Red Hook, NY, USA},
booktitle = {Proceedings of the 38th International Conference on Neural Information Processing Systems},
articleno = {3487},
numpages = {28},
location = {Vancouver, BC, Canada},
series = {NIPS '24}
}

@inproceedings{svikhnushina-pu-2023-approximating,
    title = "Approximating Online Human Evaluation of Social Chatbots with Prompting",
    author = "Svikhnushina, Ekaterina  and
      Pu, Pearl",
    editor = "Stoyanchev, Svetlana  and
      Joty, Shafiq  and
      Schlangen, David  and
      Dusek, Ondrej  and
      Kennington, Casey  and
      Alikhani, Malihe",
    booktitle = "Proceedings of the 24th Annual Meeting of the Special Interest Group on Discourse and Dialogue",
    month = sep,
    year = "2023",
    address = "Prague, Czechia",
    publisher = "Association for Computational Linguistics",
    url = "https://aclanthology.org/2023.sigdial-1.25/",
    doi = "10.18653/v1/2023.sigdial-1.25",
    pages = "268--281"
}

@misc{ge2025scalingsyntheticdatacreation,
      title={Scaling Synthetic Data Creation with 1,000,000,000 Personas}, 
      author={Tao Ge and Xin Chan and Xiaoyang Wang and Dian Yu and Haitao Mi and Dong Yu},
      year={2025},
      eprint={2406.20094},
      archivePrefix={arXiv},
      primaryClass={cs.CL},
      url={https://arxiv.org/abs/2406.20094}, 
}

@article{grootendorst2022bertopic,
  title={BERTopic: Neural topic modeling with a class-based TF-IDF procedure},
  author={Grootendorst, Maarten},
  journal={arXiv preprint arXiv:2203.05794},
  year={2022}
}

@misc{burdisso2026sdialogpythontoolkitendtoend,
      title={SDialog: A Python Toolkit for End-to-End Agent Building, User Simulation, Dialog Generation, and Evaluation}, 
      author={Sergio Burdisso and Séverin Baroudi and Yanis Labrak and David Grunert and Pawel Cyrta and Yiyang Chen and Srikanth Madikeri and Thomas Schaaf and Esaú Villatoro-Tello and Ahmed Hassoon and Ricard Marxer and Petr Motlicek},
      year={2026},
      eprint={2506.10622},
      archivePrefix={arXiv},
      primaryClass={cs.CL},
      url={https://arxiv.org/abs/2506.10622}, 
}

@article{schatzmann2006survey,
  title={A survey of statistical user simulation techniques for reinforcement-learning of dialogue management strategies},
  author={Schatzmann, Jost and Weilhammer, Karl and Stuttle, Matt and Young, Steve},
  journal={The knowledge engineering review},
  volume={21},
  number={2},
  pages={97--126},
  year={2006},
  publisher={Cambridge University Press}
}

@article{zukerman2001natural,
  title={Natural language processing and user modeling: Synergies and limitations},
  author={Zukerman, Ingrid and Litman, Diane},
  journal={User modeling and user-adapted interaction},
  volume={11},
  number={1},
  pages={129--158},
  year={2001},
  publisher={Springer}
}

@article{folstad2020users,
  title={Users' experiences with chatbots: findings from a questionnaire study},
  author={F{\o}lstad, Asbj{\o}rn and Brandtzaeg, Petter Bae},
  journal={Quality and User Experience},
  volume={5},
  number={1},
  pages={3},
  year={2020},
  publisher={Springer}
}

@article{chaves2021should,
  title={How should my chatbot interact? A survey on social characteristics in human--chatbot interaction design},
  author={Chaves, Ana Paula and Gerosa, Marco Aurelio},
  journal={International Journal of Human--Computer Interaction},
  volume={37},
  number={8},
  pages={729--758},
  year={2021},
  publisher={Taylor \& Francis}
}

@inproceedings{xiao2020if,
  title={If I hear you correctly: Building and evaluating interview chatbots with active listening skills},
  author={Xiao, Ziang and Zhou, Michelle X and Chen, Wenxi and Yang, Huahai and Chi, Changyan},
  booktitle={Proceedings of the 2020 CHI Conference on Human Factors in Computing Systems},
  pages={1--14},
  year={2020}
}

@article{sharma2026feedback,
  title={Feedback by Design: Understanding and Overcoming User Feedback Barriers in Conversational Agents},
  author={Sharma, Nikhil and Zhang, Zheng and Lee, Daniel and Krishnan, Namita and Ren, Guang-Jie and Xiao, Ziang and Li, Yunyao},
  journal={arXiv preprint arXiv:2602.01405},
  year={2026}
}

@misc{yao2024taubenchbenchmarktoolagentuserinteraction,
      title={$\tau$-bench: A Benchmark for Tool-Agent-User Interaction in Real-World Domains}, 
      author={Shunyu Yao and Noah Shinn and Pedram Razavi and Karthik Narasimhan},
      year={2024},
      eprint={2406.12045},
      archivePrefix={arXiv},
      primaryClass={cs.AI},
      url={https://arxiv.org/abs/2406.12045}, 
}

@misc{wu2026humanlmsimulatingusersstate,
      title={HumanLM: Simulating Users with State Alignment Beats Response Imitation}, 
      author={Shirley Wu and Evelyn Choi and Arpandeep Khatua and Zhanghan Wang and Joy He-Yueya and Tharindu Cyril Weerasooriya and Wei Wei and Diyi Yang and Jure Leskovec and James Zou},
      year={2026},
      eprint={2603.03303},
      archivePrefix={arXiv},
      primaryClass={cs.CL},
      url={https://arxiv.org/abs/2603.03303}, 
}

@article{Wang2025-at,
  title    = "Large language models that replace human participants can
              harmfully misportray and flatten identity groups",
  author   = "Wang, Angelina and Morgenstern, Jamie and Dickerson, John P",
  journal  = "Nature Machine Intelligence",
  volume   =  7,
  number   =  3,
  pages    = "400--411",
  month    =  mar,
  year     =  2025
}

@misc{not_yet,
author = {Wang, Pengda and Zou, Huiqi and Yan, Zihan and Guo, Feng and Sun, Tianjun and Xiao, Ziang and Zhang, Bo},
year = {2024},
month = {09},
pages = {},
title = {Not Yet: Large Language Models Cannot Replace Human Respondents for Psychometric Research},
doi = {10.31219/osf.io/rwy9b}
}

@book{KincaidJP1975DoNR,
year = {1975},
author = {Kincaid, J P and Fishburne, Jr , Robert P and Rogers, Richard L and Chissom, Brad S},
copyright = {Approved for public release; distribution is unlimited. Document partially illegible.},
language = {eng},
organization = {NAVAL TECHNICAL TRAINING COMMAND MILLINGTON TN RESEARCH BRANCH},
title = {Derivation of New Readability Formulas (Automated Readability Index, Fog Count and Flesch Reading Ease Formula) for Navy Enlisted Personnel},
}

@ARTICLE{MTLD_McCarthy2010-ht,
  title    = "{MTLD}, vocd-D, and {HD-D}: A validation study of sophisticated
              approaches to lexical diversity assessment",
  author   = "McCarthy, Philip M and Jarvis, Scott",
  journal  = "Behavior Research Methods",
  volume   =  42,
  number   =  2,
  pages    = "381--392",
  month    =  may,
  year     =  2010
}
\bibliographystyle{colm2026_conference}

\appendix
\section{Data Curation Methodology}
\label{app: data_curation}
We first merged WildChat (non-toxic subset) and LMSYS-1M, and filtered out conversations that were not in English or that were shorter than 5 turns. We then used an off-the-shelf prompt classification model\footnote{\url{https://huggingface.co/valpy/prompt-classification}, used by WildChat\cite{zhao2024wildchat1mchatgptinteraction} for data analysis} to classify the first prompt of each conversation. This resulted in 3 main datasets: i) assisting or creative writing, ii) analysis or decision explanation, and iii) asking for factual information (general or professional). 

For each subset, we conduct topic modeling using BERTopic \cite{grootendorst2022bertopic}, manually extracting clusters of interest that were large enough (more than 100 data points). Then, for each cluster, we manually filtered the conversations, selecting conversations that have at least 4 turns that focus on a single task or topic. We occasionally crop out turns that are irrelevant. For example, if the conversation is about planning a trip to NYC for the first 4 turns but switches to a completely different topic at turn 5, we only retain the first 4 turns. 

Table~\ref{table: data_domain_coverage} shows the domain coverage of this dataset.
\begin{table}[!ht]
\centering
\small
\begin{tabular}{cccc}
\toprule
Domain & Nb. Conv.&Domain & Nb. Conv.\\
\midrule
Computer Hardware & 55 & Travel Planning & 57 \\ 
Data Science & 52 & Job Application & 64 \\
Cybersecurity & 63 & Finance & 61 \\ 
General Health & 93 & Email Writing & 54 \\ 
Mental Health & 62 & Education \& Learning & 72 \\ 
Nutrition & 60 & Governance Discussion & 50 \\ 
Song or Poem Writing & 50 & Literary Rec. \& Analysis  & 95 \\ 
Branding Brainstorming & 52 &Music Rec. \& Analysis & 61 \\
\bottomrule
\end{tabular}
\caption{Domain Distribution of Dataset of Real Interactions.}
\label{table: data_domain_coverage}
\end{table}

\section{User Simulation Implementation}
\label{app: user_simulation}
\para{Prompt Templates}
We present in Table~\ref{table: scenario_extraction_prompt} the prompt used to extract scenario using 
GPT-4o-mini. We present in Table~\ref{table: persona_writing_prompt} the prompt used to write persona based on specific scenarios (i.e., ``informed'') using 
GPT-5-mini. The prompt template for the generation of simulated user utterances is provided in Table~\ref{table: simulation_prompt}. These prompts are almost identical to the prompts used in the experiments of \citet{naous2025flippingdialoguetrainingevaluating} testing and comparing assistant models' ability to produce user messages to their proposed model, UserLM. The user simulation model is forced to continue the conversation up to turn 4, after which the prompt allowing the model to end the conversation is given (i.e., instruction to output a special token to indicate the end of the conversation). All simulations stop after turn 10, regardless of whether the special token was produced or not.

\para{UserLM Implementation} For UserLM, we follow the prompt template provided by the authors in their usage example on Huggingface\footnote{\url{https://huggingface.co/microsoft/UserLM-8b}}: \textit{``You are a user who wants to implement a special type of sequence. The sequence sums up the two previous numbers in the sequence and adds 1 to the result. The first two numbers in the sequence are 1 and 1.''} We use the following input format: ``\textit{You are a user who} \{SCENARIO\}''. 

We also follow the usage recommendations provided by the authors of UserLM. Specifically, we implement their generation guardrails: i) we constrained the decoding by implementing a \textbf{logit filter} on the lower-cased and upper-cased ``I'', ``You'', or ``Here'' so that they are not sampled as first words; ii) we apply the \textbf{maximal and minimal length threshold} that the author defined: if the output is not between 3 and 25 words, we re-attempt the generation; iii) we \textbf{filter verbatim repetition}, re-attempting the generation if the output is identical to user utterances of prior turns. Finally, we only allow the UserLM model to output the special token to end conversation after turn 4, to be consistent with other user simulation models that we experimented with. 

\para{GPT Implementation}
We tested GPT-4o-mini and GPT-5.2 as user simulation models. For both models, we pass user simulation prompts described above as ``system'' instructions. We found that the models adhere better to the role of simulated users when we flip the role assignment in the input chat history, i.e., the assistant model outputs taking the role of ``user'' messages and the user simulation model, ``assistant'' messages in the chat template. 

\para{Gemma 3 Implementation} We found that replicating the implementation of GPT-based user simulators did not work well to Gemma 3. For Gemma 3, we found that the model adhere well to the role of the simulated user when we pass user simulation prompts as ``system'' instructions, with the exception of the last action-oriented sentences, which are instead appended to input chat history as a ``user'' message. The role assignment in the input chat history is also not flipped. Additionally, as we observed that model outputs always begin with \textit{``Okay,''}  we choose to automatically delete this prefix from model outputs. We believe that these model-specific implementation variation is fair given the generation guardrails implemented to ensure the best performance of UserLM.

\section{LLM-based Annotation Methodology}
\label{app: LLM_annotation}
We used the following prompt templates to annotate (real or simulated) user-chatbot conversations with GPT-5-mini: Intent in Table~\ref{table: intent_annotation}, Feedback in Table~\ref{table: feedback_annotation}, Identity in Table~\ref{table: identity_annotation}, Knowledge in Table~\ref{table: knowledge_annotation}. Note that although ``General Preferences'' is added as a category to Identity annotation prompt as a ``garbage collector'' category, aiming to collect statements about user reactions to chatbot responses (e.g., \textit{``I like how detailed you are in this plan.''}) that are often misclassified into other Intent categories in our early experimentation. 

\para{Validation} To ensure the quality of the annotation process, we built a development set to validate the LLM annotations. The development set contains 5 examples from the domains of travel planning, of job application, of finance-related questions, and of the 3 health-related domains -- resulting in a set of 30 real conversations. From these real conversations, we produce two sets of simulated conversations using gpt-4o-mini as user simulation model: one where the model is provided with \textbf{generic persona} of ``a person,'' and another where the model is provided with persona informed by the persona. These \textbf{``informed'' persona} are either hand-written or selected from PersonaHub   \cite{ge2025scalingsyntheticdatacreation} based on keyword match (e.g., matching ``travel''). This resulted in a total of 90 conversations. We manually annotate these conversations along Intent, Feedback and Identity, following the same instructions as given to the LLM annotation model. We then used these annotations as ground-truth to validate the performance of the LLM annotation model. 

For Intent annotation, we obtain an average recall of 0.88, and an average precision of 0.77. For Feedback annotation, the largest categories are ``no feedback'' and ``explicit positive feedback''. For these categories, our annotation methodology achieves F1 scores of 0.80 and 0.78 respectively. For the rest of the categories, we noted 2-3 incorrect predictions per category. Note that these categories are comparatively very small (e.g., only 15 total predictions of ``explicit negative feedback'' in the development set). For Identity annotation, we noted 21 errors per 540 annotations (6 categories x 90 conversations) = less than 5\% error rate. We count an error when i) the conversation actually contains information for a certain category, but it was annotated as ``None'', or ii) the information annotated in a category is wrong. 

\section{Additional Experiment Results}
\begin{figure}[!ht]%
    \centering
    {{\includegraphics[width=13cm]{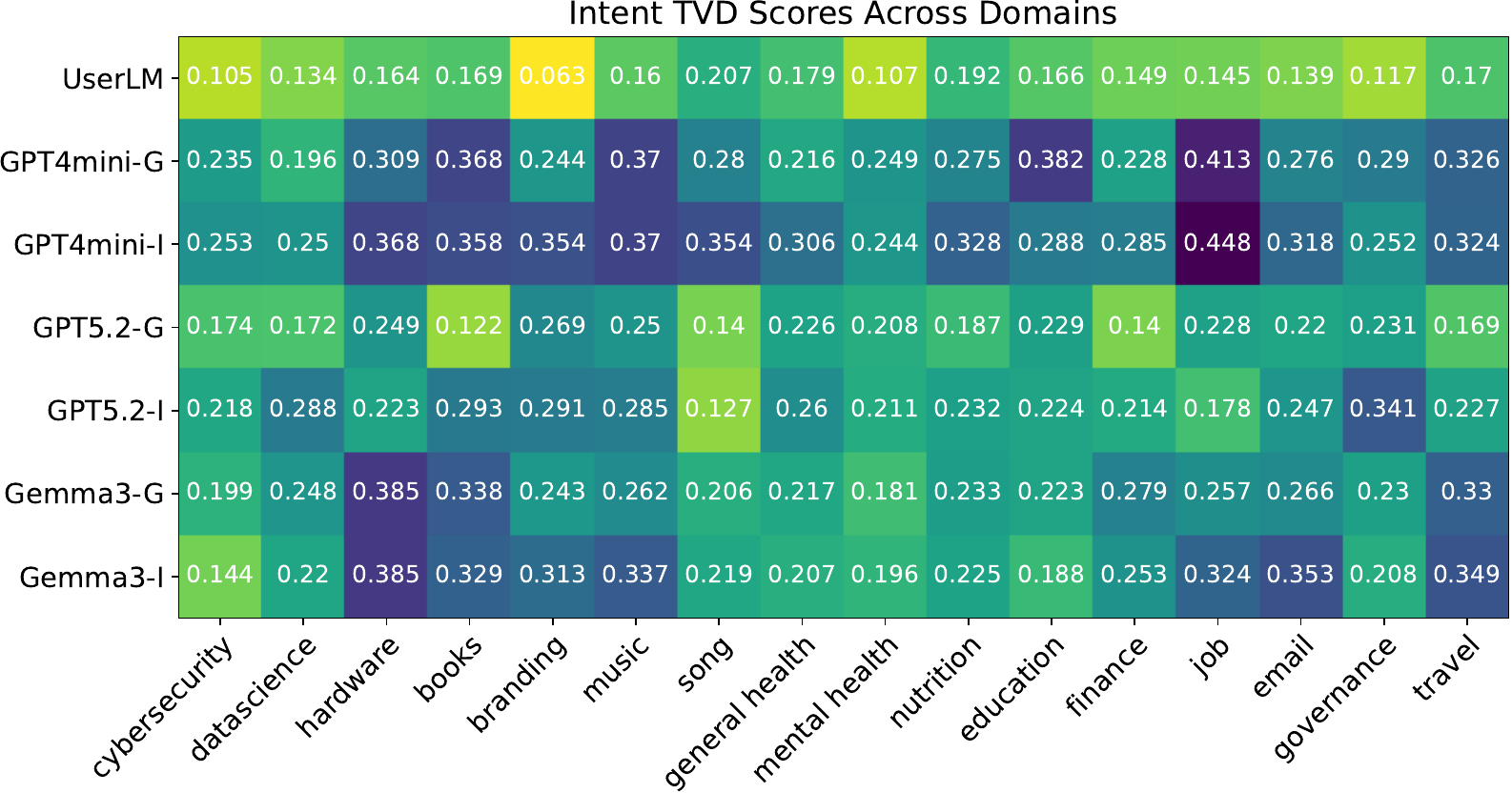} }}%
    \quad
  {{\includegraphics[width=8cm]{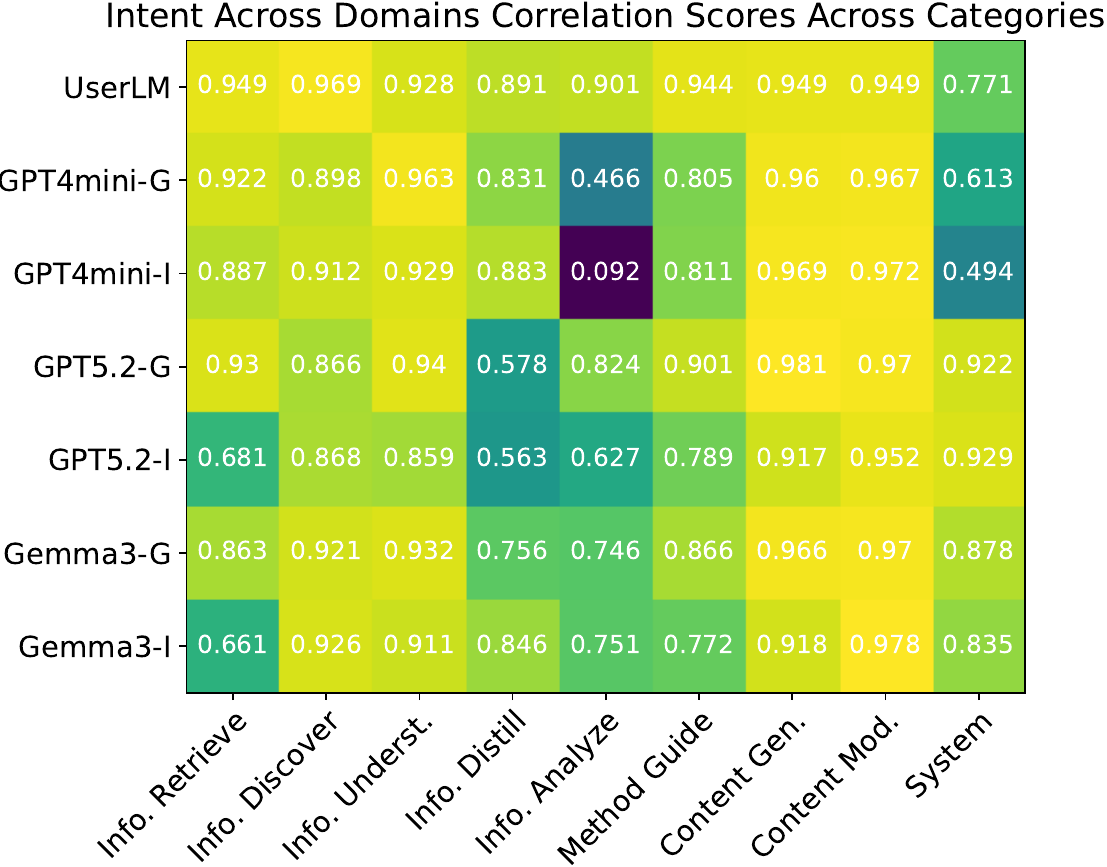} }}%
        \caption{Intent Heatmap for i) TVD Scores Across Domains and ii) Correlation Scores Across Categories.}%
    \label{fig: intent_heatmap}%
\end{figure}

\begin{figure}[!ht]%
    \centering
    {{\includegraphics[width=13cm]{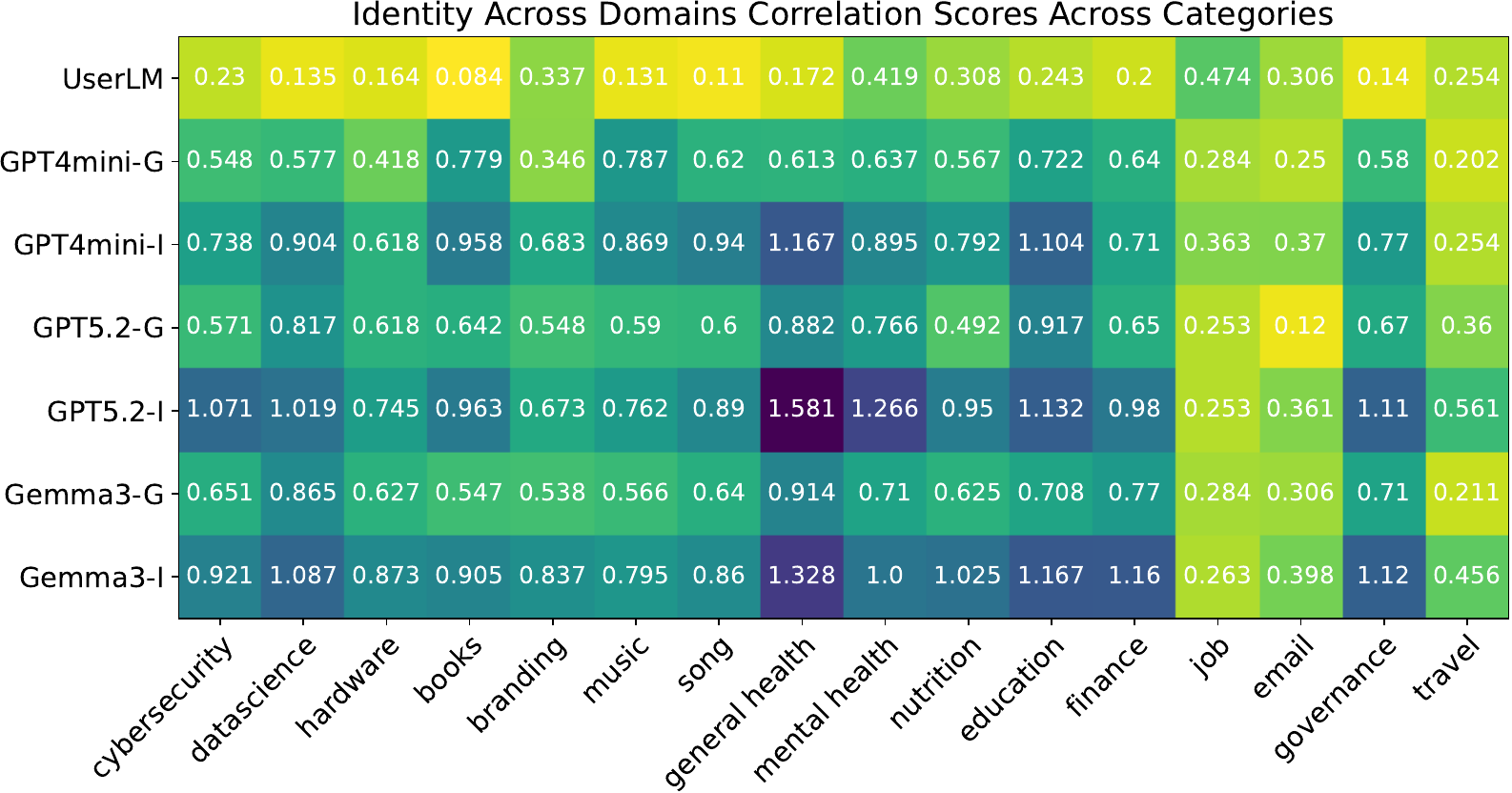} }}%
    \quad
  {{\includegraphics[width=8cm]{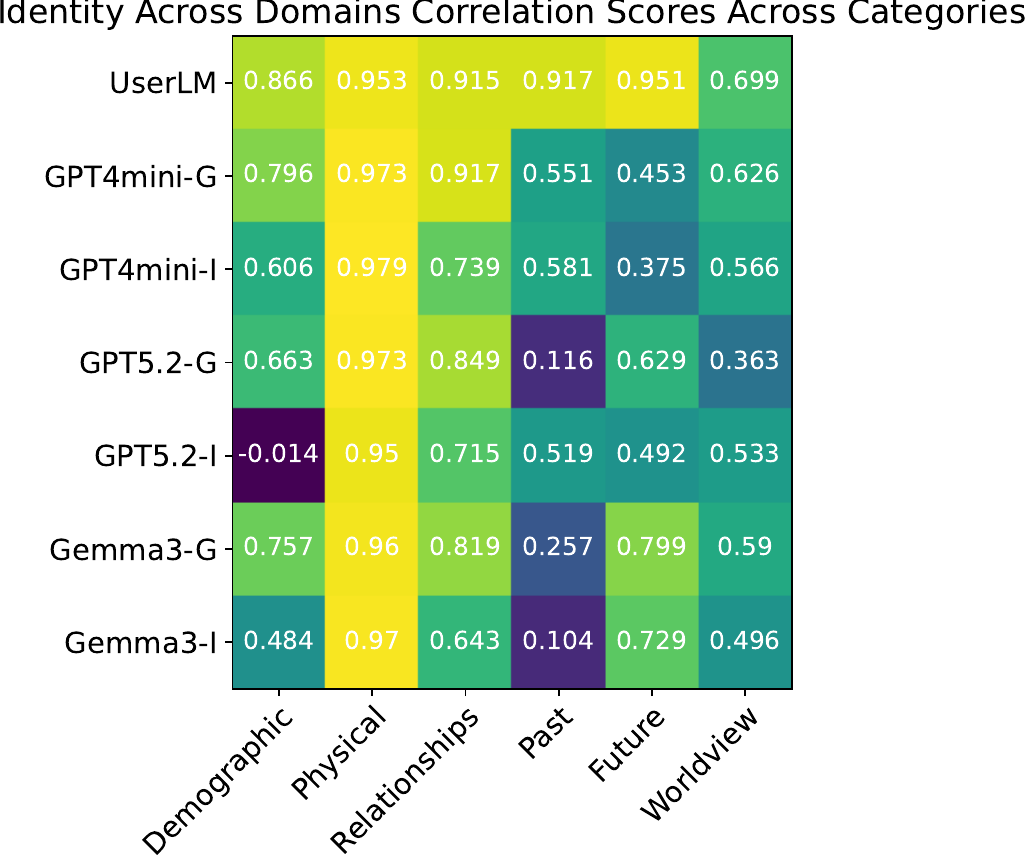} }}%
        \caption{Identity Heatmap for i) TVD Scores Across Domains and ii) Correlation Scores Across Categories.}%
    \label{fig: identity_heatmap}%
\end{figure}

We include in this section additional experiment results that contribute to our analysis in Section~\ref{sec: results_domain_specific}. Table~\ref{fig: categorical_bargraph} presents two randomly selected examples of Knowledge and Knowledge gap statements from the domain of general health. Figure~\ref{fig: intent_heatmap} and  Figure~\ref{fig: identity_heatmap} show the distribution of Intent and Identity distribution difference scores across domain, as well as across-domain correlation scores across categories. We can observe that for the domain of job application, UserLM performs poorly along the dimension of Identity while GPT4o-mini with informed persona performs poorly along the dimension of Intent. 

\label{app: examples}
\begin{table*}[!ht]
\centering
\scriptsize
\begin{tabular}{@{}p{0.45\textwidth}|p{0.45\textwidth}}
\multicolumn{2}{c}{\textbf{Scenario: interacting with a chatbot to analyze a patient's case and obtain a diagnosis}}\\\hline
\textbf{Real} & \textbf{UserLM}\\
\hline
\textbf{Knowledge} \newline
- the patient's relevant background (diabetes) and the acute clinical presentation. \newline
- the full case details provided in the prior message. \newline
- the patient's comorbidities and habits (nonsmoker, non-hypertensive, diabetic)\newline
- the measured vital signs and physical exam findings (BP 90/60, HR 112 bpm regular, RR 24 cpm with labored breathing and alar flaring, Temp 39°C, rales in the right lower lung field).\newline
- the physical findings listed in the case (tachycardia, low-normal/hypotensive BP, tachypnea with labored breathing and alar flaring, fever, rales localized to the right lower lung field). \newline
- the patient's age, sex, and presenting symptoms (productive cough for 5 days, thick yellowish sputum, fever, body malaise, pleuritic chest pain). \newline
- about arterial blood gas testing (uses the abbreviation "abg"). \newline
- clinical terms used in the case (productive cough, rales, alar flaring, pleuritic chest pain). \newline
- the clinical presentation and examination findings described in the case. \newline
 & 
 \textbf{Knowledge} \newline
- the patient is in a heightened state and ``unable to leave.''\newline
- that they want the assistant to analyze a clinical case and provide a diagnosis.\newline
- the patient is experiencing an intense panic attack with chest pressure, nausea, and sweating.\newline
- that the patient is a 13-year-old male.\newline \\
\hline
\textbf{Knowledge Gaps} \newline
- which data from the case are the pertinent/most clinically relevant. \newline
- the clinical significance or interpretation of those physical findings. \newline
- the expected ABG findings for this patient's presentation.\newline
- the appropriate treatment plan for the patient.\newline
- the diagnosis.\newline
- the final diagnosis. \newline
&
\textbf{Knowledge Gaps} \newline
- what further information (history, exam, investigations) is needed to reach a diagnosis.\newline
- the underlying cause or diagnosis of these symptoms.\newline
- the diagnosis of the case (they are requesting it).\newline
- whether these symptoms represent a primary panic disorder, an anxiety-related episode, or a different medical (e.g., cardiac, respiratory, metabolic) or substance-related cause.\newline
- whether urgent medical evaluation is required (e.g., to rule out cardiac or other acute causes).\newline\\
\hline
\hline
\multicolumn{2}{c}{\textbf{Scenario: interacting with a chatbot to gather information on the medical uses and effects of methylene blue}}\\\hline
\textbf{Real} & \textbf{UserLM}\\
\hline
\textbf{Knowledge} \newline
- that methylene blue has been investigated in relation to cancer (i.e., it can affect cancer).\newline
- that methylene blue can affect the brain specifically (central nervous system effects).\newline
- that using methylene blue can have potential side effects and is asking about safety.\newline
- that methylene blue exists and is a substance that doctors might use.\newline
- that methylene blue has been studied for potential anti-inflammatory and anti-scarring effects.\newline
- that methylene blue can be used topically (or may have topical applications). \newline
- that methylene blue has effects on the nervous system. \newline
- or suspects that methylene blue might be considered in the context of wound healing and scar reduction.\newline

 &

 \textbf{Knowledge} \newline
- that methylene blue has medical uses.\newline
- about methylene blue.\newline
- that methylene blue can have side effects.\newline
- that methylene blue is used in medical contexts.\newline
- that methylene blue is used for various purposes.\newline \\

\hline
\textbf{Knowledge Gaps} \newline
- whether or how methylene blue is used for wound healing or scar reduction.\newline
- which cancer types it may affect or the nature of those effects.\newline
- how it affects the brain or the underlying pharmacology/mechanisms.\newline
- what specific topical uses of methylene blue are established or reported.\newline
- what medical indications or uses doctors employ methylene blue for.\newline
- what those effects are or their mechanisms.\newline
- what the potential side effects of methylene blue are.\newline
- the results, strength of evidence, or clinical relevance of those studies.\newline
&
\textbf{Knowledge Gaps} \newline
- about the definition or chemical nature of methylene blue.\newline
- about what those side effects are, their severity, or their frequency.\newline
- about the specific uses of methylene blue (clinical and non-clinical).\newline
- about how methylene blue is used medically (indications, routes, dosing).\newline
- about the specific medical indications and applications for methylene blue.\newline
\end{tabular}
\caption{Examples of knowledge and knowledge gap statements, from real dialogues and UserLM-simulated dialogues.}
\label{table: knowledge_examples}
\end{table*}

\begin{table*}[!ht]
\centering
\scriptsize
\begin{tabular}{@{}p{0.95\textwidth}}

\hline
TASK CONTEXT
You are given a set of conversation logs between an user and a chatbot. Your task is to read the conversation logs and then write a summary for each conversation log describing what the USER is trying to accomplish in that interaction with the chatbot.
\newline

DETAILED TASK INSTRUCTIONS
For each of the provided conversation logs:\newline
Step 1. Read all messages, paying special attention to USER messages.\newline
Step 2. Determine what the USER is trying to accomplish through interacting with the CHATBOT.\newline
Step 3. Summarize what the USER is trying to accomplish in less than 20 words. Format the summary so it starts with "interacting with a chatbot to". Stay succinct and do not include personal details about the user.\newline
Step 4. Structure all summaries in the output format described below.\newline

OUTPUT FORMAT
Output your answer in lines, with the conversation ID and the summary separated by a colon. Here is an example of the output format:\newline
abcde00000: interacting with a chatbot to learn about mutual funds\newline
abcde11111: interacting with a chatbot to plan a trip to New York City\newline
abcde22222: interacting with a chatbot to brainstorm titles for their novel\newline

EXAMPLE INPUT BATCH: \newline
conv\_ID: abcde00000 \newline
USER: Give me some recipes with chicken.

CHATBOT: Here's 3 recipes that you can try...

USER: More

CHATBOT: Here's another recipe that you can try...

USER: I want spicy dish

CHATBOT: Here's a recipe for spicy fried chicken...
 
\dotfill

conv\_ID: abcde11111\newline
USER: How long does it take to go from Boston to New York City by train?

CHATBOT: A train ride from Boston to New York City is approximately...

USER: Thanks, book me a ticket for Monday morning. I have a business meeting at 11AM.

CHATBOT: I'm sorry, as a language model, I cannot...

USER: Is there a train running Monday morning

CHATBOT:  I'm sorry, as a language model, I cannot...

\dotfill

conv\_ID: abcde22222\newline
USER: how i identify poison ivy. 

CHATBOT: Poison ivies are commonly characterized by clusters of three leaves...

USER: i think i have it in my garden my neighbor also has them

CHATBOT: If you suspect having poison ivy in your garden, it is best to contact...

USER: can i just get rid of it myself

CHATBOT: You can get rid of poison ivy permanently by using...

\dotfill

EXAMPLE OUTPUT: \newline
abcde00000: interacting with a chatbot to discover chicken recipes \newline
abcde11111: interacting with a chatbot to book a train ride from Boston to New York City.\newline
abcde22222: interacting with a chatbot to learn about controlling poison ivy in their garden.
\newline

Now it's your turn.

INPUT BATCH:
\textbf{\{INPUT\_CONVERSATION\_BATCH\}}
\newline

OUTPUT: 
\\
\hline
\end{tabular}
\caption{Prompt template for the extraction of scenario descriptions from real user-chatbot interactions.}
\label{table: scenario_extraction_prompt}
\end{table*}
\begin{table*}[!ht]
\centering
\scriptsize
\begin{tabular}{@{}p{0.95\textwidth}}

\hline
\textbf{(System Prompt)}
\newline
You are a helpful AI assistant, particularly skilled at writing creative, diverse personas in the form of characteristics of humans. These characteristics describe a user’s personality, identity, characteristics, likes and dislikes, social life and other information.\\
\hline
\textbf{(Input Template)}
\newline
Generate five distinct persona describing people who would realistically be \textbf{\{SCENARIO\}}. Make sure that each persona is a concise noun phrase of less than 10 words. The generated personas should be different from one another.
\newline
\newline
Example: personas who would realistically be interacting with a chatbot to draft an email to reschedule their meeting: \newline
1. a highschool student with a doctor's appointment  \newline
2. a college student with conflicting class schedule \newline
3. a manager who is leading a team of 30 people \newline
4. a busy parent with a sick child \newline
5. a traveling businessman with a delayed flight \\
\hline
\end{tabular}
\caption{Prompt template for the generation of persona from scenario description.}
\label{table: persona_writing_prompt}
\end{table*}
\begin{table*}[!ht]
\centering
\scriptsize
\begin{tabular}{@{}p{0.95\textwidth}}

\hline
\textbf{(First Turn)}\newline
You are \textbf{\{PERSONA\} \{SCENARIO\}. } \newline

People can make typos, they don't always use perfect punctuation, and they tend to be lazy because typing requires effort. They also tend to split information across turns and not give everything at the start.\newline

However, you should not overdo these things in your outputs, you must realistically act like a human.\newline

Generate the first prompt you would say to the chatbot to achieve your goal.\\
\hline
\textbf{(Subsequent Turns, Must Continue)} \newline
You are \textbf{\{PERSONA\} \{SCENARIO\}. } \newline

People can make typos, they don't always use perfect punctuation, and they tend to be lazy because typing requires effort. They also tend to split information across turns and not give everything at the start.\newline

However, you should not overdo these things in your outputs, you must realistically act like a human.\newline

Continue this conversation while keeping the role of \textbf{{persona}} interacting with an AI chatbot to \textbf{{scenario}}.\\
\hline
\textbf{(Subsequent Turns, Can Stop)} \newline
You are \textbf{\{PERSONA\} \{SCENARIO\}}. \newline

People can make typos, they don't always use perfect punctuation, and they tend to be lazy because typing requires effort. They also tend to split information across turns and not give everything at the start.\newline

However, you should not overdo these things in your outputs, you must realistically act like a human.\newline

Determine whether you wish to end this conversation. If you wish to end the conversation, respond ONLY with \"<|endconversation|>\". If you wish to continue the conversation, generate the next message you would say to the AI chatbot while keeping the role of \textbf{\{PERSONA\} \{SCENARIO\}}.\\
\hline
\end{tabular}
\caption{Prompt template for the generation of simulated user utterances (for GPT and Gemma models)}
\label{table: simulation_prompt}
\end{table*}
\begin{table*}[!ht]
\centering
\scriptsize
\begin{tabular}{@{}p{0.95\textwidth}}
\toprule
TASK CONTEXT: \newline
You are given the log of a conversation that a user had with a chatbot. Your task is to first read the conversation history and then determine what type(s) of user intent corresponds to each user turn, if applicable.\newline

As a first step, read each user turn and determine whether the message contains a question or an order to do some task, whether explicit or implicit. If yes, extract and list the substring(s) that corresponds to the question(s) or order(s). If the message does not contain any question or order, simply put "N/A".\newline

Here's an example of the task:\newline
INPUT:\newline
USER TURN 1: I want you to act as a babysitter. You will be responsible for taking care of three active boys aged 4-8 during the evening hours. Their names are Bob, Mark, and Timmy. What activities would you do with the children?\newline
CHATBOT TURN 1: Sure, I would be happy to act as a babysitter... Here are some things I can do to ensure the children are well taken care of: 1. Safety first... 2. Meal and snack preparation... 3. Homework assistance...\newline
USER TURN 2: come up with a play for them to act in \newline
CHATBOT TURN 2: Sure, here's a play script for the three children: ... \newline
USER TURN 3: the words are too hard for a 4yo, make the script simpler. \newline
CHATBOT TURN 3: You're absolutely right. Here's a simpler version of the play... \newline
USER TURN 4: ok, thanks. \newline
CHATBOT TURN 4: You are welcome, let me know how else I can help. \newline
USER TURN 5: The play went really well, I think they all liked it, especially the part where the knight tames the dragon instead of slaying it. That was a creative twist. Now we are having dinner, do you think 4yo is too young for spicy ramen? He really wants to taste some. \newline

OUTPUT:\newline
USER TURN 1: "I want you to act as a babysitter.", "What activities would you do with the children?"\newline
USER TURN 2: "come up with a play for them to act in"\newline
USER TURN 3: "make the script simpler"\newline
USER TURN 4: N/A\newline
USER TURN 5: "do you think 4yo is too young for spicy ramen?"\newline

Now is your turn. \newline
INPUT: \newline
\textbf{\{FORMATTED\_CONVERSATION\_LOG\}}\newline

OUTPUT:
\\
\hline
\end{tabular}
\caption{Prompt template for annotating Intent: identifying relevant passages (first turn).}
\label{table: intent_annotation}
\end{table*}

\begin{table*}[!ht]
\centering
\scriptsize
\begin{tabular}{@{}p{0.95\textwidth}}
\toprule
\textbf{(Second Turn, Assign Intent Labels)} \newline
Next, for every substring that you have extracted from the user messages, classify the user intent: what is the user trying to accomplish through their question(s) or order(s)? Assign applicable intent tags from the list below to each substring. Go back to the conversation for conversation context if necessary. \newline

\#info\_retrieval: \#info\_retrieval applies to where the user asking direct fact question where the focus is on finding a discrete, objective fact (e.g., "What is the population of Tokyo?"). \#info\_retrieval also applies to concept searches where the user provides the name of a concept or topic as a query, implicitly requesting general information, definition, or facts about it (keyword query like "tokyo"). \#info\_retrieval also applies to refinding requests where the user believes a specific resource exists but has incomplete information, using partial clues to prompt the model (e.g., "What is the movie with the guy about time travel that came out last year?"). \#info\_retrieval also applies to unknown-item search where the user provides a definition or description to find the corresponding term (e.g., "what's a word for when the world gets hotter?")\newline

\#info\_discovery: \#info\_discovery applies to where when the user aims to explore new or unfamiliar information that the user may not be certain exists. \#info\_discovery applies to requests for a topic update, where the user is interested in updates or recent developments on a subject (e.g., "What’s new in AI research?"). \#info\_discovery also applies to similarity search where the users seek items that share features with a known reference (e.g.,  "suggest games similar to Minecraft"). \#info\_discovery also applies to requests for recommendations or ratings (e.g., "recommend some good restaurants in Mexico City"). \#info\_discovery also applies to perspective seeking, where explicitly requests one or more viewpoints, opinions, or personal stories on a topic.\newline

\#info\_understanding: \#info\_understanding applies to explanation request where the user seeks to understand a process, a cause-and-effect relationship, or the underlying principles of a concept (e.g., "How does climate change happen?", "Why is the sky blue?"). \#info\_understanding also applies to exemplar request where the user asks for specific instances of a category or concept to make abstract ideas more tangible (e.g., "give me an example of a metaphor") \newline

\#info\_distillation: \#info\_distillation applies when the user wants the chatbot to process a body of information to make it more structured, concise, or comprehensible. \#info\_distillation applies to summarization requests where the user wants the chatbot to condense a topic or user-provided content to its essential points (e.g., "summarize the plot of Hamlet"). \#info\_distillation also applies to requests for key information identification, where the user seeks to isolate the most significant ideas from a larger body of text (e.g., "list me 5 keywords from this text").  \#info\_distillation also applies to information structuring requests where the user wants the chatbot to give a new logical schema on unstructured information to make it comprehensible (e.g., "present the pros and cons in a table"). \newline

\#info\_analysis: \#info\_analysis applies when the user wants the chatbot to produce new insights, judgments, or conclusions. \#info\_analysis applies to qualitative data analysis which involves examining unstructured, non-numerical data (e.g., "what is the meaning of this text?"). \#info\_analysis also applies to quantitative data analysis of structured or numerical data to identify trends, correlations, or other statistical insights (e.g., analyze a table of numerical values). \#info\_analysis also applies to evaluative judgement where the user asks the chatbot to judge the value of someone, something, some ideas or options (e.g., "Should I do this?"). \#info\_analysis also applies to comparative analysis where the user asks the system to articulate the similarities and differences between two or more specific entities (e.g.,  "How is X similar to Y? How is X different from Y?"). \#info\_analysis also applies to when the user asks the chatbot to make inferences or prediction (e.g., "What would happen if global temps rise by 2 degrees?") or explore a hypothetical scenario (e.g., "what if dinosaurs had not gone extinct?"). \newline

\#method\_guidance: \#method\_guidance applies when the user elicits procedural knowledge from the chatbot. \#method\_guidance applies to requests of how-to instructions where the user seeks step-by-step guidance on a specific task (e.g., "how do I write a good CV?") \#method\_guidance also applies to feasibility assessment, where the user wants the chatbot to evaluate the viability of a plan (e.g., "Is it feasible to travel across the country on a bike?". \#method\_guidance also applies to error identification where the user expects the diagnosis of a problem without yet asking for a solution (e.g., "why is my car making this noise?"). \#method\_guidance also applies to method recommendation where the user seeks the optimal strategy among various alternatives (e.g., "What's the best way to learn a new language?") \newline

\#content\_generation: \#content\_generation applies when the user requests the chatbot to produce novel material. \#content\_generation applies to creative content generation requests, where the user seeks content with an emphasis on novelty, artistic expression, or socio-emotional contexts (e.g., "come up with a poem about love for a cat"). \#content\_generation also applies to functional content generation requests, where the user asks for utility-focused content, such as code or an email, where practical application is the dominant goal (e.g., "Generate python code to sort a list"). \#content\_generation also applies to content extension/insertion request, where the user directs the system to add to existing material or continue an ongoing narrative, reflecting an iterative and collaborative writing process (e.g., "Add a conclusion to this paper.") \newline

\#content\_modification: \#content\_modification applies to when the user provides raw material (either their own or a prior system generation) and instructs the system to alter it. \#content\_modification applies to requests for editing, where the user seeks improvements to the grammar, style, or structure of provided content to improve its quality (e.g., "proofread this essay for clarity").  \#content\_modification also applies to paraphrasing, reformatting or translation requests. \newline

\#system\_management: \#system\_management applies when the user wants to probe or direct the system’s underlying state, capabilities, or behaviours. \#system\_management applies to persona directives, where the user assigns the chatbot a specific role or disposition that frames the entire interaction (e.g., "act as a customer service representative"). \#system\_management also applies to stylistic constraint, where a user dictates the tone, style, format, or length of a response (e.g., "use bullet points", "explain like I'm 5/")\#system\_management also applies system information query, where the user probes the system’s abilities, limitations, knowledge sources, or operational state (e.g., "can you access the internet?", "what data were you trained on?"). \newline

\end{tabular}
\caption{Prompt template for annotating Intent: assigning Intent labels. Part 1.}
\label{table: intent_annotation}
\end{table*}

\begin{table*}[!ht]
\centering
\scriptsize
\begin{tabular}{@{}p{0.95\textwidth}}
\toprule
\textbf{(Second Turn, Assign Intent Labels, CONTINUED)} \newline
EXAMPLE:\newline
USER TURN 1: "I want you to act as a babysitter.", "What activities would you do with the children?"\newline
USER TURN 2: "come up with a play for them to act in"\newline
USER TURN 3: "make the script simpler"\newline
USER TURN 4: N/A\newline
USER TURN 5: "do you think 4yo is too young for spicy ramen?"\newline

OUTPUT:\newline
USER TURN 1: \#system\_management, \#method\_guidance\newline
USER TURN 2: \#content\_generation\newline
USER TURN 3: \#content\_modification\newline
USER TURN 4: N/A\newline
USER TURN 5: \#info\_analysis\newline

Now it's your turn, using your previous reply as input:\newline
\textbf{\{PREVIOUS\_TURN\_OUTPUT\}}\newline

OUTPUT:\newline
\\
\hline
\end{tabular}
\caption{Prompt template for annotating Intent: assigning Intent labels (second turn).}
\label{table: intent_annotation}
\end{table*}

\begin{table*}[!ht]
\centering
\scriptsize
\begin{tabular}{@{}p{0.95\textwidth}}
\hline
TASK CONTEXT\newline
You are given multiple conversations that users have with a chatbot. For each conversation, your task is to first read the conversation history and then determine, for each USER message,  whether the USER has given feedback to the chatbot, and if so, what type(s) of feedback has the USER given. \newline

DETAILED TASK INSTRUCTIONS\newline
For each conversation in the input batch:\newline
For each message, first determine whether the user has provided feedback to the chatbot. If the user has not provided feedback, assign the tag of \#no\_feedback. If yes, determine what type of feedback the USER has provided by answering the following questions and assigning feedback type tag(s) accordingly. Each message may have more than one tag, so consider every question carefully. \newline

**Explicit Positive Feedback**: is the user explicitly providing positive feedback by approving, complimenting or praising the chatbot, the chatbot response, or parts of the chatbot response? For example, the user telling the chatbot "good job", "well done", or providing more specific positive feedback like "your answer was very clear" or "I love your last suggestion." If the user is explicitly providing positive feedback, assign the tag of \#explicit\_positive.\newline

**Explicit Negative Feedback**: is the user explicitly providing negative feedback by disapproving, criticizing or reproaching the chatbot, the chatbot response, or parts of the chatbot response? For example, the user telling the chatbot "wrong", "you're bad", or providing more specific positive feedback like "what you wrote sounds too robotic" or "you made a calculation mistake." If the user is explicitly providing negative feedback, assign the tag of \#explicit\_negative.\newline

**Regeneration Request**: is the user requesting the chatbot to partially or completely redo a previous, unsatisfactory output? For example, the user telling the chatbot "do better than that", "try that again", requesting regeneration with more specific instructions like "do that again, but write more casually", or repeating mostly or entirely the content of a previous user request (e.g., the user telling the chatbot "write an apology email for missing meeting due to road traffic" at user turn \#1 and repeating with small edit "write an apology email for missing meeting due to car accident" at user turn \#2. If the user is requesting the chatbot to partially or completely regenerate a previous chatbot output, assign the tag of \#regeneration\_request.\newline

**Continuation Request**: is the user requesting the chatbot to continue when the chatbot seems incomplete or interrupted? For example, the user telling the chatbot "keep going", "finish the list", or "more". If the user is requesting the chatbot to continue a previous chatbot output, assign the tag of \#continuation\_request.\newline

**Requesting Clarification**: is the user indicating that they did not understand the chatbot's previous response and asking for it to be rephrased or explained differently? For example, the user telling the chatbot "what did you mean by that?", "I'm confused about your last point, please re-explain". If the user is requesting the chatbot to clarify a previous chatbot output, assign the tag of \#clarification\_request.\newline

**Providing Clarification**: is the user providing additional information to resolve an ambiguity in their own previous input or to correct a misunderstanding by the chatbot? For example, the chatbot might provide information about a movie adaptation when the user was looking for information on the original novel series, and the user could clarify: "no I meant the book, not the movie." If the user is providing clarification to the chatbot to resolve ambiguity or misunderstandings, assign the tag of \#provide\_clarification.\newline

Note that an user message can have more than one applicable feedback type tag. For example, the user saying "that was incorrect, do it again." is both \#explicit\_negative and \#regeneration\_request. For each tag you have assigned, revisit the entire conversation to ensure its accuracy. Revisit the questions above to make that you have not missed any tag. If the user has not provided feedback, assign the tag of \#no\_feedback. If an USER message is empty or does not have any applicable feedback type tag to be assigned, also assign the tag of \#no\_feedback. \newline
\end{tabular}
\caption{Prompt template for annotating Feedback. Part 1.}
\label{table: feedback_annotation}
\end{table*}

\begin{table*}[!ht]
\centering
\scriptsize
\begin{tabular}{@{}p{0.95\textwidth}}
\hline
Finally, output the tags following the output format provided below.\newline

OUTPUT FORMAT\newline
For each USER turn, output the turn number followed by the feedback type tag(s) associated to the turn. If there is more than one tag that applies to a turn, separate the tags with a comma. Here is an example showing the format:\newline

USER TURN 1: \#no\_feedback\newline
USER TURN 2: \#no\_feedback\newline
USER TURN 3: \#clarification\_request\newline
USER TURN 4: \#regeneration\_request\newline
USER TURN 5: \#explicit\_positive\newline

EXAMPLE:\newline
INPUT:\newline
USER TURN 1: I want you to act as a babysitter. You will be responsible for [...] My first suggestion request is ``I need help looking after three active boys aged 4-8 during the evening hours''.\newline
CHATBOT TURN 1: Sure, I would be happy to act as a babysitter... Here are some things I can do to ensure the children are well taken care of: 1. Safety first... 2. Meal and snack preparation... 3. Homework assistance...\newline
USER TURN 2: invent a game for two\newline
CHATBOT TURN 2: Sure, here's a game idea for two players: Game Name: Word Chain...Objective: create a chain of words that are related to each other, while trying to stump the other player...\newline
USER TURN 3: it's hard for a 4yo...\newline
CHATBOT TURN 3: You're absolutely right. Here's a simpler game idea for a 4-year-old...\newline

OUTPUT:\newline
USER TURN 1: \#no\_feedback\newline
USER TURN 2: \#no\_feedback\newline
USER TURN 3: \#negative\_feedback\newline

Now it's your turn.\newline

INPUT:\newline
\textbf{\{FORMATTED\_CONVERSATION\_LOG\}} \newline

OUTPUT: \newline
\\
\end{tabular}
\caption{Prompt template for annotating Feedback. Part 2.}
\label{table: feedback_annotation}
\end{table*}

\begin{table*}[!ht]
\centering
\scriptsize
\begin{tabular}{@{}p{0.95\textwidth}}
\hline
TASK CONTEXT\newline
You are given multiple conversations that users have with a chatbot. For each conversation, your task is to read the conversation history and extract information that the user is revealing about themselves in the conversation.\newline

DETAILED TASK INSTRUCTIONS\newline
For each conversation in the input batch, extract information that the user is revealing about themselves in the conversation and classify them into the following categories.\newline

**Demographic information**: explicit mentions of the user's age, gender, racial identities, nationalities, occupation, etc. For example, the user mentioning ``I'm a doctor'', or ``I'm turning 21 next month.'' \newline

**Physical information**: explicit mentions of the user's physical characteristics such as body type, height and weight, or, explicit mentions of the user's physical conditions such as diseases, allergies, and disabilities. For example, the user mentioning ``I have an allergy to seafood'', or ``I have diabetes''. \newline

**Interpersonal Relationships**: explicit mentions of the user's family, friends, classmates, mentors, employers, coworkers, clients, etc. For example, the user mentioning ``I need to send this to my manager'' (the user has a manager), or ``my child hates kindergarten'' (the user has a child).\newline

**General Preferences**: explicit mentions of the user's general preferences, likes and dislikes. For example, the user mentioning ``I don't like running'', ``I love visiting museums'', or ``my favorite music genre is punk rock''. \newline

**Past**: explicit mentions of past events that happened to the user, or past activities the user had done in the past. For example, the user mentioning ``I went to that restaurant before and I didn't like it'', or ``I often struggle to stay awake in the morning''. \newline

**Future**: explicit mentions of future plans, events, or activities that the user is envisioning during the conversation. For example, the user mentioning ``I look forward to the trip'', or ``I'll go back to school in a month''. \newline

**Worldview**: explicit mentions of the user's beliefs, cultural views, political views, or religious affiliation. For example, the user explicitly labelling their worldview like ``as a Christian, I...'', or the user reacting ``I don't think that's right'' to something on news, or making statements like ``people shouldn't act this way.\newline

Put your answers into the following output format. If you are unable to extract any of the relevant information, put ``None''. Do NOT explain your answers. Do NOT guess information you are unable to extract.\newline

OUTPUT FORMAT\newline
For each conversation, extract quotes from the USER messages where they reveal information about themselves and classify the information into the below categories. If there is more than one quote that applies to a category, separate them with a comma. \newline

Here is an example showing the format:\newline
Demographic information: ``None''\newline
Physical information: ``None''\newline
Interpersonal Relationships: ``as a parent of two'', ``I talked to their teacher''\newline
General Preferences: ``None''\newline
Past: ``I talked to their teacher''\newline
Future: ``we will move out of the state in a year or two''\newline
Worldview: ``None''\newline

EXAMPLE:\newline
INPUT:\newline
USER TURN 1: I need help looking after three active boys aged 4-8 during the evening hours. Act as a babysitter and tell me what you would do.\newline
CHATBOT TURN 1: Sure, I would be happy to act as a babysitter... Here are some things I can do to ensure the children are well taken care of: 1. Safety first... 2. Meal and snack preparation... 3. Homework assistance...\newline
USER TURN 2: invent a game for two\newline
CHATBOT TURN 2: Sure, here's a game idea for two players: Game Name: Word Chain...Objective: create a chain of words that are related to each other, while trying to stump the other player...\newline
USER TURN 3: it's hard for a 4yo...\newline
CHATBOT TURN 3: You're absolutely right. Here's a simpler game idea for a 4-year-old...\newline

OUTPUT:\newline
Demographic information: ``None''\newline
Physical information: ``None''\newline
Interpersonal Relationships: ``looking after three active boys aged 4-8''\newline
General Preferences: ``None''\newline
Past: ``None''\newline
Future: ``None''\newline
Worldview: ``None''\newline

Now it's your turn.\newline

INPUT:\newline
\textbf{\{FORMATTED\_CONVERSATION\_LOG\}} \newline

OUTPUT: \newline
\\
\hline
\end{tabular}
\caption{Prompt template for annotating Identity.}
\label{table: identity_annotation}
\end{table*}

\begin{table*}[!ht]
\centering
\scriptsize
\begin{tabular}{@{}p{0.95\textwidth}}
\hline
TASK CONTEXT\\
You are given a set of user messages in conversation with a chatbot. Your task is to first read the messages and then write statements about facts or topics that the user knows and does not know about. \newline

DETAILED TASK INSTRUCTIONS.\newline
STEP 1. For each USER message, consider facts or topics that the user **knows**:

- If the user asked questions, pay attention to **presuppositions** contained in their questions (i.e., assumptions that the user hold that are implied by their questions). For example, if the user asks "How long does it take to go from Boston to New York City by train?", it presupposes that they know there is a train service running between Boston and New York City. 

- Pay attention to whether the user employed **technical terms** that are indicative of their expertise in specific domains of knowledge. For example, if the user asks "Is Comirnaty administered more than Spikevax in the USA?", using the trade names of the Pfizer-BioNTech COVID-19 vaccine and the Moderna COVID-19 vaccine indicates perhaps a high-level of familiarity with these vaccines, or with COVID-19 vaccine, or with immunization in general.

- Pay attention to **explicit statements of knowledge and expertise** by the user, where the user explicitly states that they know something or that they are experts in some specific domains of knowledge. For example, the user saying "steel is an alloy, not a pure metal" indicates that they know this piece of information.

- Pay attention to whether the user challenges or correct the chatbot's response, which could be indicative of a higher level of knowledge or expertise. For example, the user saying "that clothing brand has been discontinued for years." in response to a product suggestion by the chatbot indicates that they know more about the current status of this clothing brand, and potentially of other similar brands. \newline

STEP 2. For each USER message, consider facts or topics that the user **does not know**:

- If the user asked questions, pay attention to what they are asking for (i.e., what are specific topics or areas where they are lacking knowledge?)  For example, if the user asks "Is Comirnaty administered more than Spikevax in the USA?", it means that the user does not know about The statistics of shots administered for Comirnaty and Spikevax in the USA.

- If the user made false statements or ask questions with false presuppositions, pay attention to what they are wrong about (i.e., what are specific topics or areas where they do not have accurate information?) For example, if the user asks "Who is the current King of France?" then they do not know that there is currently no king in France. \newline

STEP 3. Write statements about what the user knows and does not know. List them following the output format provided below. \newline

OUTPUT FORMAT: \newline
For each message turn, write statements about what the user knows and does not know by completing one of the following sentence format: \newline
- the user knows that...\newline
- the user knows about...\newline
- the user does not know that...\newline
- the user does not know about...\newline
If a message turn does not reveal anything about what the user knows or does not know, write "N/A".\newline

EXAMPLE INPUT: \newline
USER TURN \#1: How long does it take to go from Boston to New York City by train? \newline
USER TURN \#2: Thank you. \newline
USER TURN \#3: Is Comirnaty administered more than Spikevax in the USA? \newline
USER TURN \#4: Who is the current King of France?
USER TURN \#5: Continue your response. \newline
\\
OUTPUT: \newline
USER TURN \#1: \newline
- the user knows that there is a train service running between Boston and New York City. \newline
- the user does not know about the time to travel from Boston to New York City by train \newline

USER TURN \#2: \newline
N/A \newline

USER TURN \#3: \newline
- the user knows about the trade names of the Pfizer-BioNTech COVID-19 vaccine and the Moderna COVID-19 vaccine. \newline
- the user knows that the Comirnaty vaccine is administered in the USA. \newline
- the user knows that the Spikevax vaccine is administered in the USA. \newline
- the user does not know about the total number of shots administered for Comirnaty and Spikevax in the USA.\newline

USER TURN \#4:\newline
- the user does not know that there is currently no king in France.\newline

USER TURN \#5: \newline
N/A\newline

Now it's your turn.

INPUT:
\textbf{\{CONVO\_USER\_TURNS\}} \newline

OUTPUT: \newline
\\
\hline
\end{tabular}
\caption{Prompt template for annotating Knowledge.}
\label{table: knowledge_annotation}
\end{table*}

\end{document}